\documentclass[times,art10,twocolumn,latex8]{article}

\usepackage{iccv}
\usepackage{microtype}
\usepackage{graphicx}
\usepackage{booktabs} 
\usepackage{amsmath}
\usepackage{multicol}
\usepackage{multirow}
\usepackage{bbm}
\usepackage{float}
\usepackage{caption}
\usepackage{xspace}
\usepackage[separate-uncertainty=true]{siunitx}
\usepackage{etoolbox}
\usepackage[dvipsnames]{xcolor}
\usepackage{dblfloatfix}
\usepackage{placeins}
\usepackage{lipsum}
\usepackage{times}
\usepackage{subcaption}
\usepackage{wrapfig}
\usepackage{amssymb}

\graphicspath{{gfx/}}

\usepackage{url}
\usepackage[pagebackref=true,breaklinks=true,letterpaper=true,colorlinks,bookmarks=false]{hyperref}

\iccvfinalcopy 

\begin{document}

\title{SLAMP: Stochastic Latent Appearance and Motion Prediction}


\author{Adil Kaan Akan$^1$
\qquad
Erkut Erdem$^2$
\qquad 
Aykut Erdem$^1$
\qquad 
Fatma G\"uney$^1$\\
$^1$ Ko\c{c} University Is Bank AI Center, Istanbul, Turkey\\
$^2$ Hacettepe University Computer Vision Lab, Ankara, Turkey\\
{\tt\small \{kakan20,aerdem,fguney\}@ku.edu.tr} \qquad {\tt\small erkut@cs.hacettepe.edu.tr}\\
{\tt\small\textbf{\url{https://kuis-ai.github.io/slamp}}}
}

\maketitle


\newcommand{\Perp}{\perp\!\!\! \perp}
\newcommand{\bK}{\mathbf{K}}
\newcommand{\bX}{\mathbf{X}}
\newcommand{\bY}{\mathbf{Y}}
\newcommand{\bk}{\mathbf{k}}
\newcommand{\bx}{\mathbf{x}}
\newcommand{\by}{\mathbf{y}}
\newcommand{\bhy}{\hat{\mathbf{y}}}
\newcommand{\bty}{\tilde{\mathbf{y}}}
\newcommand{\bG}{\mathbf{G}}
\newcommand{\bI}{\mathbf{I}}
\newcommand{\bg}{\mathbf{g}}
\newcommand{\bS}{\mathbf{S}}
\newcommand{\bs}{\mathbf{s}}
\newcommand{\bM}{\mathbf{M}}
\newcommand{\bw}{\mathbf{w}}
\newcommand{\eye}{\mathbf{I}}
\newcommand{\bU}{\mathbf{U}}
\newcommand{\bV}{\mathbf{V}}
\newcommand{\bW}{\mathbf{W}}
\newcommand{\bn}{\mathbf{n}}
\newcommand{\bv}{\mathbf{v}}
\newcommand{\bwv}{\mathbf{wv}}
\newcommand{\bq}{\mathbf{q}}
\newcommand{\bR}{\mathbf{R}}
\newcommand{\bi}{\mathbf{i}}
\newcommand{\bj}{\mathbf{j}}
\newcommand{\bp}{\mathbf{p}}
\newcommand{\bt}{\mathbf{t}}
\newcommand{\bJ}{\mathbf{J}}
\newcommand{\bu}{\mathbf{u}}
\newcommand{\bB}{\mathbf{B}}
\newcommand{\bD}{\mathbf{D}}
\newcommand{\bz}{\mathbf{z}}
\newcommand{\bP}{\mathbf{P}}
\newcommand{\bC}{\mathbf{C}}
\newcommand{\bA}{\mathbf{A}}
\newcommand{\bZ}{\mathbf{Z}}
\newcommand{\bff}{\mathbf{f}}
\newcommand{\bF}{\mathbf{F}}
\newcommand{\bo}{\mathbf{o}}
\newcommand{\bO}{\mathbf{O}}
\newcommand{\bc}{\mathbf{c}}
\newcommand{\bm}{\mathbf{m}}
\newcommand{\bT}{\mathbf{T}}
\newcommand{\bQ}{\mathbf{Q}}
\newcommand{\bL}{\mathbf{L}}
\newcommand{\bl}{\mathbf{l}}
\newcommand{\ba}{\mathbf{a}}
\newcommand{\bE}{\mathbf{E}}
\newcommand{\bH}{\mathbf{H}}
\newcommand{\bd}{\mathbf{d}}
\newcommand{\br}{\mathbf{r}}
\newcommand{\be}{\mathbf{e}}
\newcommand{\bb}{\mathbf{b}}
\newcommand{\bh}{\mathbf{h}}
\newcommand{\bhh}{\hat{\mathbf{h}}}
\newcommand{\btheta}{\boldsymbol{\theta}}
\newcommand{\bTheta}{\boldsymbol{\Theta}}
\newcommand{\bpi}{\boldsymbol{\pi}}
\newcommand{\bphi}{\boldsymbol{\phi}}
\newcommand{\bpsi}{\boldsymbol{\psi}}
\newcommand{\bPhi}{\boldsymbol{\Phi}}
\newcommand{\bmu}{\boldsymbol{\mu}}
\newcommand{\bsigma}{\boldsymbol{\sigma}}
\newcommand{\bSigma}{\boldsymbol{\Sigma}}
\newcommand{\bGamma}{\boldsymbol{\Gamma}}
\newcommand{\bbeta}{\boldsymbol{\beta}}
\newcommand{\bomega}{\boldsymbol{\omega}}
\newcommand{\blambda}{\boldsymbol{\lambda}}
\newcommand{\bLambda}{\boldsymbol{\Lambda}}
\newcommand{\bkappa}{\boldsymbol{\kappa}}
\newcommand{\btau}{\boldsymbol{\tau}}
\newcommand{\balpha}{\boldsymbol{\alpha}}
\newcommand{\nR}{\mathbb{R}}
\newcommand{\nN}{\mathbb{N}}
\newcommand{\nL}{\mathbb{L}}
\newcommand{\nF}{\mathbb{F}}
\newcommand{\cN}{\mathcal{N}}
\newcommand{\cM}{\mathcal{M}}
\newcommand{\cR}{\mathcal{R}}
\newcommand{\cB}{\mathcal{B}}
\newcommand{\cL}{\mathcal{L}}
\newcommand{\cH}{\mathcal{H}}
\newcommand{\cS}{\mathcal{S}}
\newcommand{\cT}{\mathcal{T}}
\newcommand{\cO}{\mathcal{O}}
\newcommand{\cC}{\mathcal{C}}
\newcommand{\cP}{\mathcal{P}}
\newcommand{\cE}{\mathcal{E}}
\newcommand{\cI}{\mathcal{I}}
\newcommand{\cF}{\mathcal{F}}
\newcommand{\cK}{\mathcal{K}}
\newcommand{\cY}{\mathcal{Y}}
\newcommand{\cX}{\mathcal{X}}
\def\bgamma{\boldsymbol\gamma}

\newcommand{\specialcell}[2][c]{%
  \begin{tabular}[#1]{@{}c@{}}#2\end{tabular}}

\newcommand{\figref}[1]{\Fig~\ref{#1}}
\newcommand{\secref}[1]{Section~\ref{#1}}
\newcommand{\algref}[1]{Algorithm~\ref{#1}}
\newcommand{\eqnref}[1]{Eq.~\eqref{#1}}
\newcommand{\tabref}[1]{Table~\ref{#1}}

\newcommand{\rulesep}{\unskip\ \vrule\ }



\newcommand{\KLD}[2]{D_{\mathrm{KL}} \left( \left. \left. #1 \right|\right| #2 \right) }
\newcommand{\KLDD}[2]{D_{\mathrm{KL}} ( #1~||~#2 ) }

\renewcommand{\b}{\ensuremath{\mathbf}}

\def\mc{\mathcal}
\def\mb{\mathbf}

\newcommand{\T}{^{\raisemath{-1pt}{\mathsf{T}}}}

\makeatletter
\DeclareRobustCommand\onedot{\futurelet\@let@token\@onedot}
\def\@onedot{\ifx\@let@token.\else.\null\fi\xspace}
\def\eg{e.g\onedot} \def\Eg{E.g\onedot}
\def\ie{i.e\onedot} \def\Ie{I.e\onedot}
\def\cf{cf\onedot} \def\Cf{Cf\onedot}
\def\etc{etc\onedot} \def\vs{vs\onedot}
\def\wrt{wrt\onedot}
\def\dof{d.o.f\onedot}
\def\etal{et~al\onedot} \def\iid{i.i.d\onedot}
\def\Fig{Fig\onedot} \def\Eqn{Eqn\onedot} \def\Sec{Sec\onedot} \def\Alg{Alg\onedot}
\makeatother

\newcommand{\xdownarrow}[1]{%
  {\left\downarrow\vbox to #1{}\right.\kern-\nulldelimiterspace}
}

\newcommand{\xuparrow}[1]{%
  {\left\uparrow\vbox to #1{}\right.\kern-\nulldelimiterspace}
}

\renewcommand\UrlFont{\color{blue}\rmfamily}

\newcommand*\rot{\rotatebox{90}}
\newcommand{\boldparagraph}[1]{\vspace{0.2cm}\noindent{\bf #1:} }

\newcommand{\ka}[1]{ \noindent {\color{blue} {#1}} }
\newcommand{\aykut}[1]{ \noindent {\color{green} {\bf Aykut:} {#1}} }
\newcommand{\ee}[1]{ \noindent {\color{cyan} {\bf Erkut:} {#1}} }

\begin{abstract}
Motion is an important cue for video prediction and often utilized by separating video content into static and dynamic components. Most of the previous work utilizing motion is deterministic but there are stochastic methods that can model the inherent uncertainty of the future.
Existing stochastic models either do not reason about motion explicitly or make limiting assumptions about the static part. In this paper, we reason about appearance and motion in the video stochastically by predicting the future based on the motion history.
Explicit reasoning about motion without history already reaches the performance of current stochastic models. The motion history further improves the results by allowing to predict consistent dynamics several frames into the future. Our model performs comparably to the state-of-the-art models on the generic video prediction datasets, however, significantly outperforms them on two challenging real-world autonomous driving datasets with complex motion and dynamic background.
\end{abstract}
\section{Introduction}
\label{sec:intro}

Videos contain visual information enriched by motion. Motion is a useful cue for reasoning about human activities or interactions between objects in a video. Given a few initial frames of a video, our goal is to predict several frames into the future, as realistically as possible.
By looking at a few frames, humans can predict what will happen next. Surprisingly, they can even attribute semantic meanings to random dots and recognize motion patterns \cite{Johansson1973}. This shows the importance of motion to infer the dynamics of the video and to predict the future frames.

\begin{figure}[h!]
\centering
\begin{subfigure}[b]{0.49\textwidth}
   \includegraphics[width=1\linewidth]{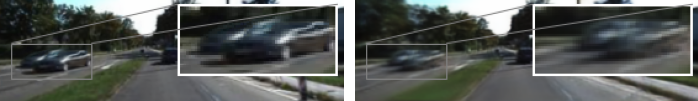}
\end{subfigure}
\begin{subfigure}[b]{0.49\textwidth}
   \includegraphics[width=1\linewidth]{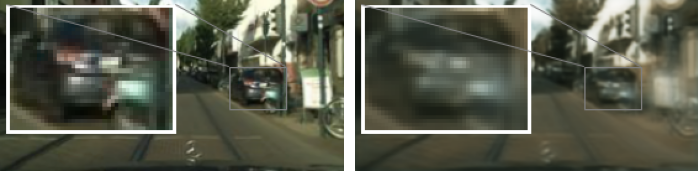}
\end{subfigure}
\caption{Comparison of the first prediction frames (11th) SLAMP~(\textbf{left}) \vs state-of-the-art method, SRVP~\cite{Franceschi2020ICML}~(\textbf{right}) on KITTI~\cite{Geiger2013IJRR}~(\textbf{top}) and Cityscapes~\cite{Cordts2016CVPR}~(\textbf{bottom}) datasets. Our method can predict both foreground and background objects better than SRVP. Full sequence predictions can be seen in \figref{fig:full_seq_teaser_city} and \ref{fig:full_seq_teaser_kitti}.}
\label{fig:teaser}
\vspace{-5mm}
\end{figure}

Motion cues have been heavily utilized for future frame prediction in computer vision.
A common approach is to factorize the video into static and dynamic components \cite{Walker2015ICCV, Liu2017ICCV, Lu2017CVPR, Fan2019AAAI, Gao2019ICCV, Lotter2017ICLR, Brabandere2016NeurIPS, Vondrick2017CVPR}. 
First, most of the previous methods are deterministic and fail to model the uncertainty of the future.
Second, motion is typically interpreted as local changes from one frame to the next. However, changes in motion follow certain patterns when observed over some time interval. 
Consider scenarios where objects move with near-constant velocity, or humans repeating atomic actions in videos.
Regularities in motion can be very informative for future frame prediction.
In this work, we propose to explicitly model the change in motion, or \emph{the motion history}, for predicting future frames.

Stochastic methods have been proposed to model the inherent uncertainty of the future in videos. Earlier methods encode the dynamics of the video in stochastic latent variables which are decoded to future frames in a deterministic way~\cite{Denton2018ICML}. We first assume that both appearance and motion are encoded in the stochastic latent variables and decode them separately into appearance and motion predictions in a deterministic way. Inspired by the previous deterministic methods~\cite{Finn2016NeurIPS, Liu2017ICCV, Gao2019ICCV}, we also estimate a mask relating the two. Both appearance and motion decoders are expected to predict the full frame but they might fail due to occlusions around motion boundaries. Intuitively, we predict a probabilistic mask from the results of the appearance and motion decoders to combine them into a more accurate final prediction. Our model learns to use motion cues in the dynamic parts and relies on appearance in the occluded regions.

The proposed stochastic model with deterministic decoders cannot fully utilize the motion history, even when motion is explicitly decoded.
In this work, we propose a model to recognize regularities in motion and remember them in the motion history to improve future frame predictions.
We factorize stochastic latent variables as static and dynamic components to model the motion history in addition to the appearance history. We learn two separate distributions representing appearance and motion and then decode static and dynamic parts from the respective ones. 

Our model outperforms all the previous work and performs comparably to the state-of-the-art method, SRVP, \cite{Franceschi2020ICML} without any limiting assumptions on the changes in the static component on the generic video prediction datasets, MNIST, KTH and BAIR. However, our model outperforms all the previous work, including SRVP, on two challenging real-world autonomous driving datasets with dynamic background and complex object motion.
\begin{figure}
\includegraphics[width=0.5\textwidth]{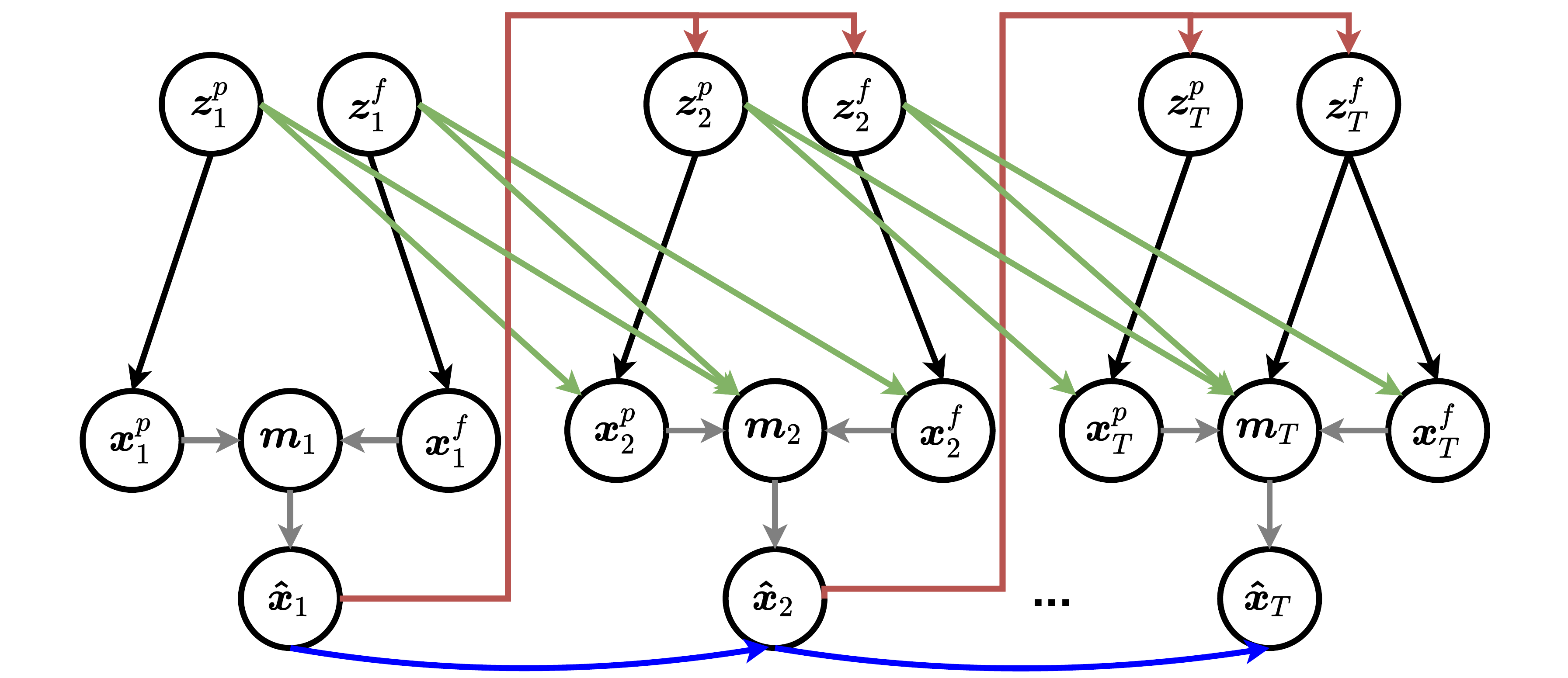}
\caption{\textbf{Generative Model of SLAMP.} The graphical model shows the generation process of SLAMP with motion history. There are two separate latent variables for appearance $\bz^p_{t}$ and motion $\bz^f_{t}$ generating frames $\bx^p_{t}$ and $\bx^f_{t}$ (black).
Information is propagated between time-steps through the recurrence between frame predictions (\textcolor{blue}{blue}), corresponding latent variables (\textcolor{LimeGreen}{green}), and from frame predictions to latent variables (\textcolor{OrangeRed}{red}).
The final prediction $\hat{\bx}_{t}$ is a weighted combination of the $\bx^p_{t}$ and $\bx^f_{t}$ according to the mask $\bm(\bx^p_{t},\bx^f_{t})$.
Note that predictions at a time-step depend on all of the previous time-steps recurrently, but only the connections between consecutive ones are shown for clarity.}
\label{fig:graphical_model}
\vspace{-5mm}
\end{figure}

\section{Related Work}
\label{sec:rw}

\vspace{-2.5mm}
\boldparagraph{Appearance-Motion Decomposition}
The previous work explored motion cues for video generation either explicitly with optical flow \cite{Walker2015ICCV, Walker2016ECCV, Liang2017ICCV, Liu2017ICCV, Lu2017CVPR, Fan2019AAAI, Gao2019ICCV} or implicitly with temporal differences \cite{Lotter2017ICLR} or pixel-level transformations \cite{Brabandere2016NeurIPS, Vondrick2017CVPR}. There are some common factors among these methods such as using recurrent models \cite{Shi2015NeurIPS, Lotter2017ICLR, Fan2019AAAI}, specific processing of dynamic parts \cite{Brabandere2016NeurIPS, Liang2017ICCV, Fan2019AAAI, Gao2019ICCV}, utilizing a mask \cite{Finn2016NeurIPS, Liu2017ICCV, Gao2019ICCV}, and adversarial training \cite{Vondrick2017CVPR, Lu2017CVPR}. We also use recurrent models, predict a mask, and separately process motion, but in a stochastic way.

The previous work which explored motion for video generation are mostly deterministic, therefore failing to capture uncertainty of the future. There are a couple of attempts to learn multiple future trajectories from a single image with a conditional variational autoencoder \cite{Walker2016ECCV} or to capture motion uncertainty with a probabilistic motion encoder \cite{Liang2017ICCV}. The latter work uses separate decoders for flow and frame similar to our approach, however, predicts them only from the latent vector. We incorporate information from previous frames with additional modelling of the motion history.

\boldparagraph{Stochastic Video Generation}
SV2P \cite{Babaeizadeh2018ICLR} and SVG \cite{Denton2018ICML} are the first to model the stochasticity in video sequences using latent variables. The input from past frames are encoded in a posterior distribution to generate the future frames. In a stochastic framework, learning is performed by maximizing the likelihood of the observed data and minimizing the distance of the posterior distribution to a prior distribution, either fixed \cite{Babaeizadeh2018ICLR} or learned from previous frames \cite{Denton2018ICML}. Since time-variance in the model is proven crucial by the previous work, we sample a latent variable at every time step \cite{Denton2018ICML}. Sampled random variables are fed to a frame predictor, modelled recurrently using an LSTM. We model appearance and motion distributions separately and train two frame predictors for static and dynamic parts.

Typically, each distribution, including the prior and the posterior, is modeled with a recurrent model such as an LSTM. Villegas et al. \cite{Villegas2019NeurIPS} replace the linear LSTMs with convolutional ones at the cost of increasing the number of parameters. Castrejon et al. \cite{Castrejon2019ICCV} introduce a hierarchical representation to model latent variables at different scales, by introducing additional complexity. Lee et al. \cite{Lee2018ARXIV} incorporate an adversarial loss into the stochastic framework to generate sharper images, at the cost of less diverse results.
Our model with linear LSTMs can generate diverse and sharp-looking results without any adversarial losses, by incorporating motion information successfully into the stochastic framework.
Recent methods model dynamics of the keypoints to avoid errors in pixel space and achieve stable learning \cite{Minderer2019NeurIPS}. This offers an interesting solution for videos with static background and moving foreground objects that can be represented with keypoints. Our model can generalize to videos with changing background without needing keypoints to represent objects.

Optical flow has been used before in future prediction~\cite{Li2018ECCV, Lu2017CVPR}. Li \etal~\cite{Li2018ECCV} generate future frames from a still image by using optical flow generated by an off-the-shelf model, whereas we compute flow as part of prediction. Lu \etal~\cite{Lu2017CVPR} use optical flow for video extrapolation and interpolation without modeling stochasticity. Long-term video extrapolation results show the limitation of this work in terms of predicting future due to relatively small motion magnitudes considered in extrapolation. Differently from flow, Xue \etal~\cite{Xue2016NeurIPS} model the motion as image differences using cross convolutions. 

\boldparagraph{State-Space Models} 
Stochastic models are typically auto-regressive, \ie the next frame is predicted based on the frames generated by the model. As opposed to interleaving process of auto-regressive models, state-space models separate the frame generation from the modelling of dynamics \cite{Gregor2019ICLR}. State-of-the-art method SRVP \cite{Franceschi2020ICML} proposes a state-space model for video generation with deterministic state transitions representing residual change between the frames. This way, dynamics are modelled with latent state variables which are independent of previously generated frames. Although independent latent states are computationally appealing, they cannot model the motion history of the video. In addition, content variable designed to model static background cannot handle changes in the background. We can generate long sequences with complex motion patterns by explicitly modelling the motion history without any limiting assumptions about the dynamics of the background.

\section{Methodology}
\label{sec:method}
\begin{figure*}[ht]
\centering
\includegraphics[width=\textwidth]{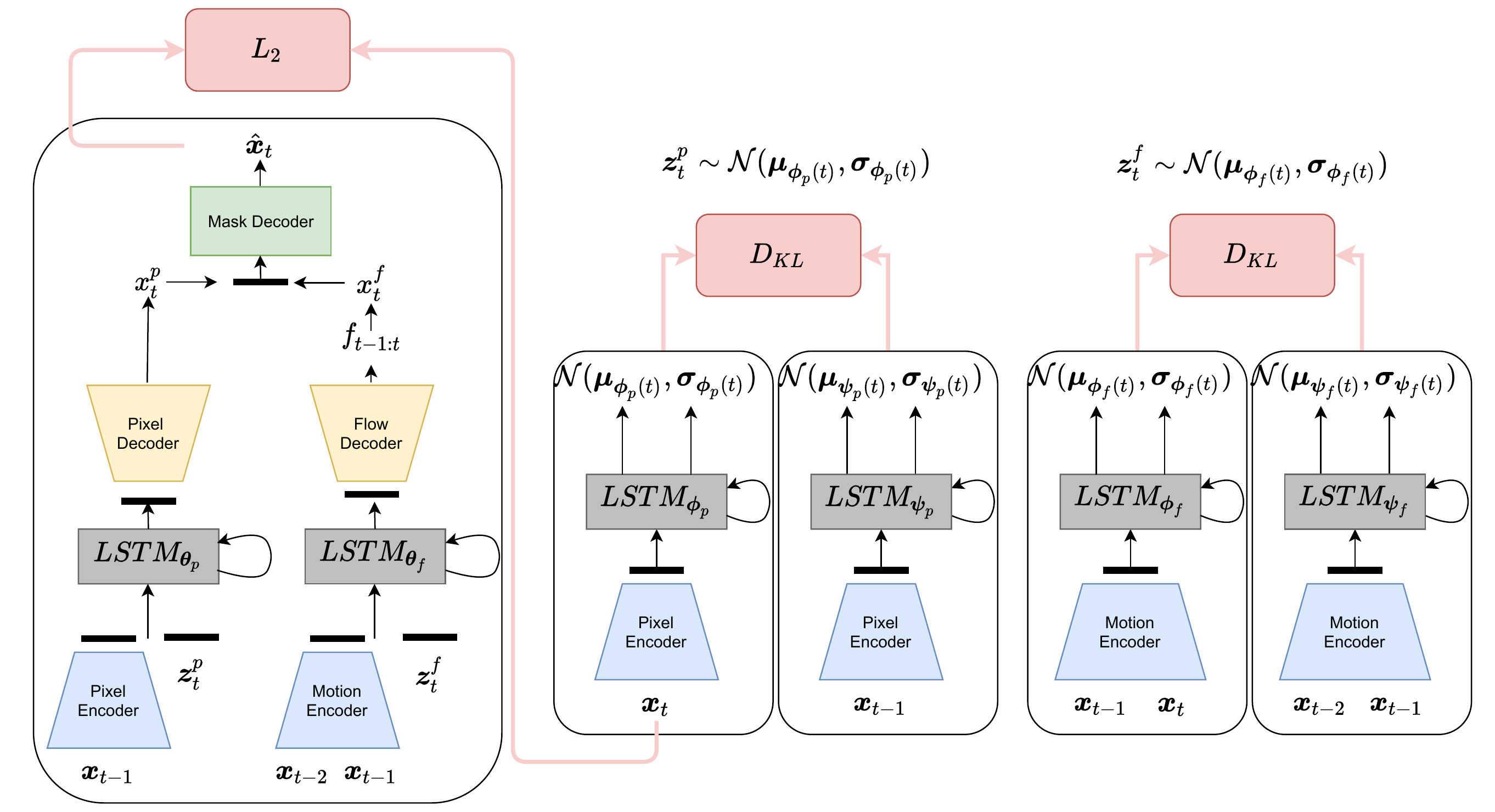}
\caption{\textbf{SLAMP.} This figure shows the components of our SLAMP model including the prediction model, inference and learned prior models for pixel and then flow from left to right. Observations $\bx_t$ are mapped to the latent space by using a pixel encoder for appearance on each frame and and a motion encoder for motion between consecutive frames. The blue boxes show encoders, yellow and green ones decoders, gray ones recurrent posterior, prior, and predictor models, and lastly red ones show loss functions during training. Note that $L_2$ loss is applied three times for appearance prediction $\bx_t^p$, motion prediction $\bx_t^f$, and the combination of the two $\hat{\bx}_t$ according to the mask prediction $\bm(\bx_t^p,\bx_t^f)$. We only show L2 loss between the actual frame $\bx_t$ and the final predicted frame $\hat{\bx}_{t}$ in the figure. For inference, only the prediction model and learned prior models are used.}
\label{fig:model_motion_prior}
\vspace{-5mm}
\end{figure*}

\subsection{Stochastic Video Prediction}
\label{sec:svg}
Given the previous frames $\bx_{1:t-1}$ until time $t$, our goal is to predict the target frame $\bx_t$. For that purpose, we assume that we have access to the target frame $\bx_t$ during training and use it to capture the dynamics of the video in stochastic latent variables $\bz_t$. By learning to approximate the distribution over $\bz_t$, we can decode the future frame $\bx_t$ from $\bz_t$ and the previous frames $\bx_{1:t-1}$ at test time.

Using all the frames including the target frame, we compute a posterior distribution $q_{\bphi}(\bz_t \vert \bx_{1:t})$ and sample a latent variable $\bz_t$ from this distribution at each time step. 
The stochastic process of the video is captured by the latent variable $\bz_t$. In other words, it should contain information accumulated over the previous frames rather than only condensing the information on the current frame. This is achieved by encouraging $q_{\bphi}(\bz_t \vert \bx_{1:t})$ to be close to a prior distribution $p(\bz)$ in terms of KL-divergence. 
The prior can be sampled from a fixed Gaussian 

at each time step or can be learned from previous frames up to the target frame $p_{\bpsi}(\bz_t \vert \bx_{1:t-1})$. We prefer the latter one as it is shown to work better by learning a prior that varies across time \cite{Denton2018ICML}.

The target frame $\bx_t$ is predicted based on the previous frames $\bx_{1:t-1}$ and the latent vectors $\bz_{1:t}$.

In practice, we only use the latest frame $\bx_{t-1}$ and the latent vector $\bz_t$ as input and dependencies from further previous frames are propagated with a recurrent model. The output of the frame predictor $\bg_t$
 
contains the information required to decode $\bx_t$. 

Typically, $\bg_t$ is decoded to a fixed-variance Gaussian distribution whose mean is the predicted target~frame $\hat{\bx}_t$~\cite{Denton2018ICML}.

\subsection{SLAMP}
\label{sec:slamp}

We call the predicted target frame, appearance prediction $\bx_t^p$ in the pixel space.
In addition to $\bx_t^p$, we also estimate optical flow $\bff_{t-1:t}$ from the previous frame $t-1$ to the target frame $t$. The flow $\bff_{t-1:t}$ represents the motion of the pixels from the previous frame to the target frame. We reconstruct the target frame $\bx_t^f$ from the estimated optical flow via differentiable warping \cite{Jaderberg2015NeurIPS}. Finally, we estimate a mask $\bm(\bx_t^p,\bx_t^f)$ from the two frame estimations to combine them into the final estimation $\hat{\bx}_t$:
\begin{equation}
    \label{eq:combined_x}
   \hat{\bx}_t = \bm(\bx_t^p,\bx_t^f) \odot \bx_t^p + (\mathbf{1} - \bm(\bx_t^p,\bx_t^f)) \odot \bx_t^f
\end{equation}
where $\odot$ denotes element-wise Hadamard product and $\bx_t^f$ is the result of warping the source frame to the target frame according to the estimated flow field $\bff_{t-1:t}$.
Especially in the dynamic parts with moving objects, the target frame can be reconstructed accurately using motion information. In the occluded regions where motion is unreliable, the model learns to rely on the appearance prediction. The mask prediction learns a weighting between the appearance and the motion predictions for combining them. 

We call this model SLAMP-Baseline because it is limited in the sense that it only considers the motion with respect to the previous frame while decoding the output. In SLAMP, we extend the stochasticity in the appearance space to the motion space as well. This way, we can model appearance changes and motion patterns in the video explicitly and make better predictions of future. \figref{fig:model_motion_prior} shows an illustration of SLAMP (see Appendix \secref{sec:model_figs} for \mbox{SLAMP-Baseline}).

In order to represent appearance and motion, we compute two separate posterior distributions $q_{\bphi_p}(\bz_t^p \vert \bx_{1:t})$ and $q_{\bphi_f}(\bz_t^f \vert \bx_{1:t})$, respectively. We sample two latent variables $\bz_t^p$ and $\bz_t^f$ from these distributions in the pixel space and the flow space. This allows a decomposition of the video into static and dynamic components. Intuitively, we expect the dynamic component to focus on changes and the static to what remains constant from the previous frames to the target frame. 
If the background is moving according to a camera motion, the static component can model the change in the background assuming that it remains constant throughout the video, \eg ego-motion of a car.

\boldparagraph{The Motion History} The latent variable $\bz_t^f$ should contain motion information accumulated over the previous frames rather than local temporal changes between the last frame and the target frame. We achieve this by encouraging $q_{\bphi_f}(\bz_t^f \vert \bx_{1:t})$ to be close to a prior distribution in terms of KL-divergence. 
Similar to \cite{Denton2018ICML}, we learn the motion prior conditioned on previous frames up to the target frame: $p_{\bpsi_f}(\bz_t^f \vert \bx_{1:t-1})$. We repeat the same for the static part represented by $\bz_t^p$ with posterior $q_{\bphi_p}(\bz_t^p \vert \bx_{1:t})$
and the learned prior $p_{\bpsi_p}(\bz_t^p \vert \bx_{1:t-1})$.

\subsection{Variational Inference}
\label{sec:inf}
For our basic formulation (SLAMP-Baseline), the derivation of the loss function is straightforward and provided in Appendix \secref{sec:deriv}. For SLAMP, the conditional joint probability corresponding to the graphical model in \figref{fig:graphical_model} is:

\vspace{-6mm}
\begin{align}
    p(\bx_{1:T}) = \prod_{t=1}^{T} &~~p(\bx_t \lvert \bx_{1:t-1}, \bz_t^p, \bz_t^f)  \\
    &~~p(\bz_{t}^{p} \lvert \bx_{1:t-1}, \bz_{t-1}^{p})
    ~~p(\bz_{t}^{f} \lvert \bx_{1:t-1}, \bz_{t-1}^{f}) \nonumber
    \vspace{-3mm}
\end{align}
The true distribution over the latent variables $\bz_{t}^{p}$ and $\bz_{t}^{f}$ is intractable. We train time-dependent inference networks $q_{\bphi_p}(\bz_{t}^{p} \lvert \bx_{1:T})$ and $q_{\bphi_f}(\bz_{t}^{f} \lvert \bx_{1:T})$ to approximate the true distribution with conditional Gaussian distributions.
In order to optimize the likelihood of $p(\bx_{1:T})$, we need to infer latent variables $\bz_{t}^{p}$ and $\bz_{t}^{f}$, which correspond to uncertainty of static and dynamic parts in future frames, respectively. We use a variational inference model to infer the latent variables.

Since $\bz_{t}^{p}$ and $\bz_{t}^{f}$ are independent across time, we can decompose Kullback-Leibler terms into individual time steps. We train the model by optimizing the variational lower bound (see Appendix \secref{sec:deriv} for the derivation):

\begin{align}
    \label{eq:elbo}
    & \mathrm{log}\ p_{\btheta}(\bx) \geq \mathcal{L}_{\btheta,\bphi_{p}, \bphi_{f},\bpsi_{p},\bpsi_{f}}
    (\bx_{1:T}) \\
    & ={} 
    \begin{aligned}[t]
        \sum\limits_{t}
            \mathbbm{E}_{\substack{\bz_{1:t}^p \sim q_{\bphi_p} \\
                                   \bz_{1:t}^f \sim q_{\bphi_f}}}
                &~\mathrm{log}\ p_{\btheta}(\bx_t \lvert \bx_{1:t-1}, \bz_{1:t}^{p}, \bz_{1:t}^{f}) \nonumber \\
                & - \beta \Big[
                \KLDD{q(\bz^{p}_t \lvert \bx_{1:t})}{p(\bz^{p}_t \lvert \bx_{1:t-1})}  \nonumber \\
      & ~~~~~ +  \KLDD{q(\bz^{f}_t \lvert \bx_{1:t})}{p(\bz^{f}_t \lvert \bx_{1:t-1})} 
                 \Big] \nonumber
    \end{aligned}
\end{align}

The likelihood $p_{\btheta}$, can be interpreted as an $L_2$ penalty between the actual frame $\bx_t$ and the estimation $\hat{\bx}_t$ as defined in \eqref{eq:combined_x}. We apply the $L_2$ loss to the predictions of appearance and motion components as well.

The posterior terms for uncertainty are estimated as an expectation over $q_{\bphi_p}(\mathbf{z}_t^p \lvert \mathbf{x}_{1:t})$, $q_{\bphi_f}(\mathbf{z}_t^f \lvert \mathbf{x}_{1:t})$.
As in \cite{Denton2018ICML}, we also learn the prior distributions from the previous frames up to the target frame as 
$p_{\bpsi_p}(\mathbf{z}_t^p \lvert \mathbf{x}_{1:t-1})$, $p_{\bpsi_f}(\mathbf{z}_t^f \lvert \mathbf{x}_{1:t-1})$.
We train the model using the re-parameterization trick \cite{Kingma2014ICLR}.
We classically choose the posteriors to be factorized Gaussian so that all the KL divergences can be computed analytically.

\subsection{Architecture}
\label{sec:arch}
We encode the frames with a feed-forward convolutional architecture to obtain appearance features at each time-step. In SLAMP, we also encode consecutive frame pairs into a feature vector representing the motion between them. 
We then train linear LSTMs to infer posterior and prior distributions 
at each time-step from encoded appearance and motion features.

Stochastic video prediction model with a learned prior \cite{Denton2018ICML} is a special case of our baseline model with a single pixel decoder, we also add motion and mask decoders.
Next, we describe the steps of the generation process for the dynamic part.

At each time step, we encode $\bx_{t-1}$ and $\bx_{t}$ into $\bh_{t}^f$, representing the motion from the previous frame to the target frame.
The posterior LSTM is updated based on the~$\bh_{t}^f$:

\begin{eqnarray}
    \bh_t^f &=& \mathrm{MotionEnc}(\bx_{t-1}, \bx_t) \\
    \bmu_{\bphi_f(t)}, \bsigma_{\bphi_f(t)} &=& \mathrm{LSTM}_{\bphi_f}(\bh_t^f) \nonumber
    \vspace{-5mm}
\end{eqnarray}
For the prior, we use the motion representation $\bh_{t-1}^f$ from the previous time step, \ie the motion from the frame $t-2$ to the frame $t-1$, to update the prior LSTM: 
\begin{eqnarray}
    \bh_{t-1}^f &=& \mathrm{MotionEnc}(\bx_{t-2}, \bx_{t-1}) \\
    \bmu_{\bpsi_f(t)}, \bsigma_{\bpsi_f(t)} &=& \mathrm{LSTM}_{\bpsi_f}(\bh_{t-1}^f) \nonumber
    \vspace{-5mm}
\end{eqnarray}
At the first time-step where there is no previous motion, we assume zero-motion by estimating the motion from the previous frame to itself.

The predictor LSTMs are updated according to encoded features and sampled latent variables:
\begin{eqnarray}
    \bg_t^f &=& \mathrm{LSTM}_{\btheta_f}(\bh_{t-1}^f, \bz_t^f) \\
    \bmu_{\btheta_f} &=& \mathrm{FlowDec}(\bg_t^f) \nonumber
\end{eqnarray}
There is a difference between the train time and inference time in terms of the distribution the latent variables are sampled from. At train time, latent variables are sampled from the posterior distribution. At test time, they are sampled from the posterior for the conditioning frames and from the prior for the following frames.
The output of the predictor LSTMs are decoded into appearance and motion predictions separately and combined into the final prediction using the mask prediction (\eqnref{eq:combined_x}).

\section{Experiments}
\label{sec:exp}
\begin{figure}[ht]
    \centering
    \includegraphics[width=0.47\textwidth]{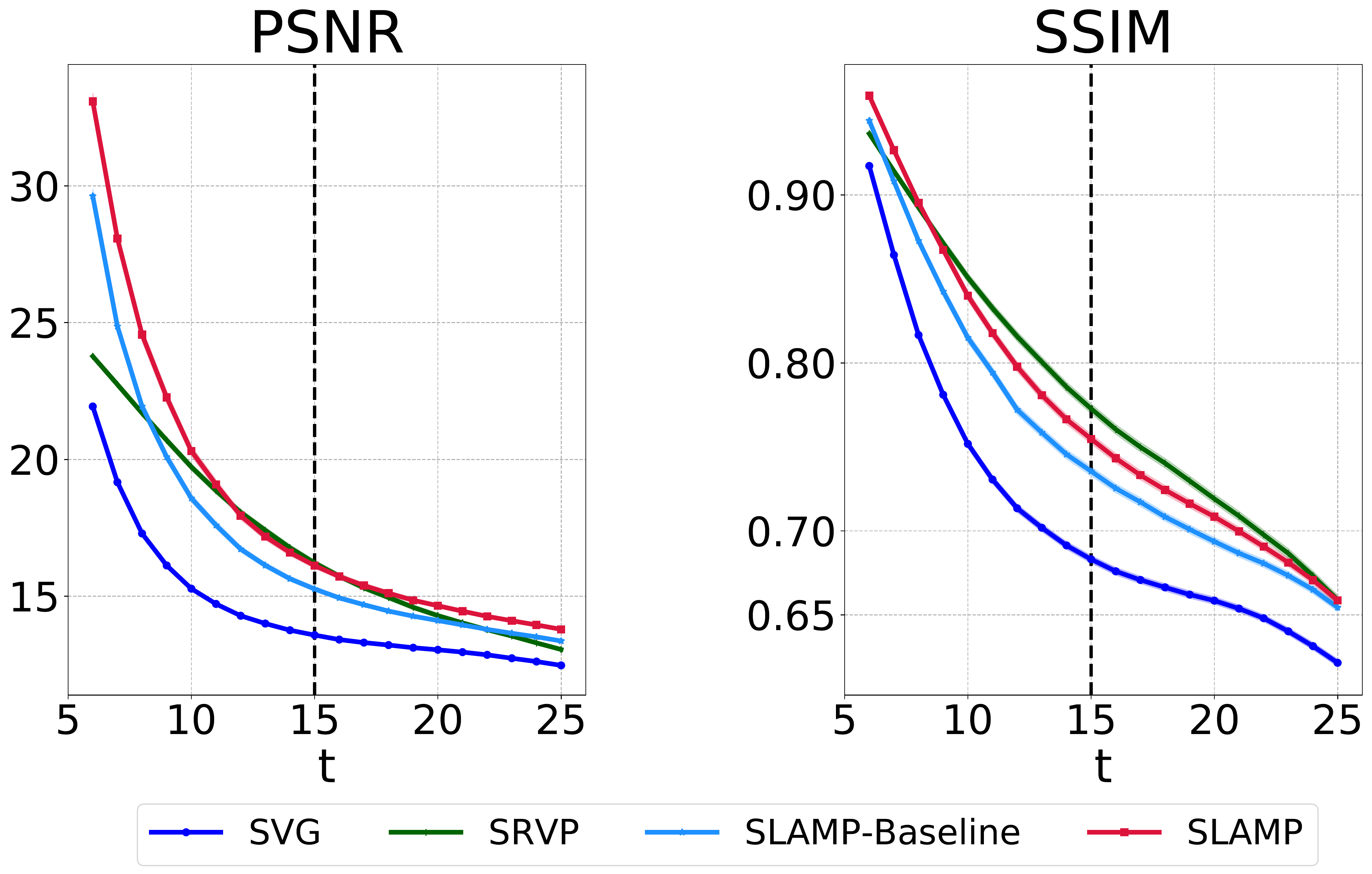}
    \caption{\textbf{Quantitative Results on MNIST.} This figure compares SLAMP to SLAMP-Baseline, SVG~\cite{Denton2018ICML}, and SRVP~\cite{Franceschi2020ICML} on MNIST in terms of PSNR~(\textbf{left}) and SSIM~(\textbf{right}). SLAMP clearly outperforms our baseline model and SVG, and performs comparably to SRVP. Vertical bars mark the length of the training sequences.}
    \label{fig:results_mnist}
    \vspace{-3.3mm}
\end{figure}
\begin{figure*}[ht]
    \centering
    \includegraphics[width=\textwidth]{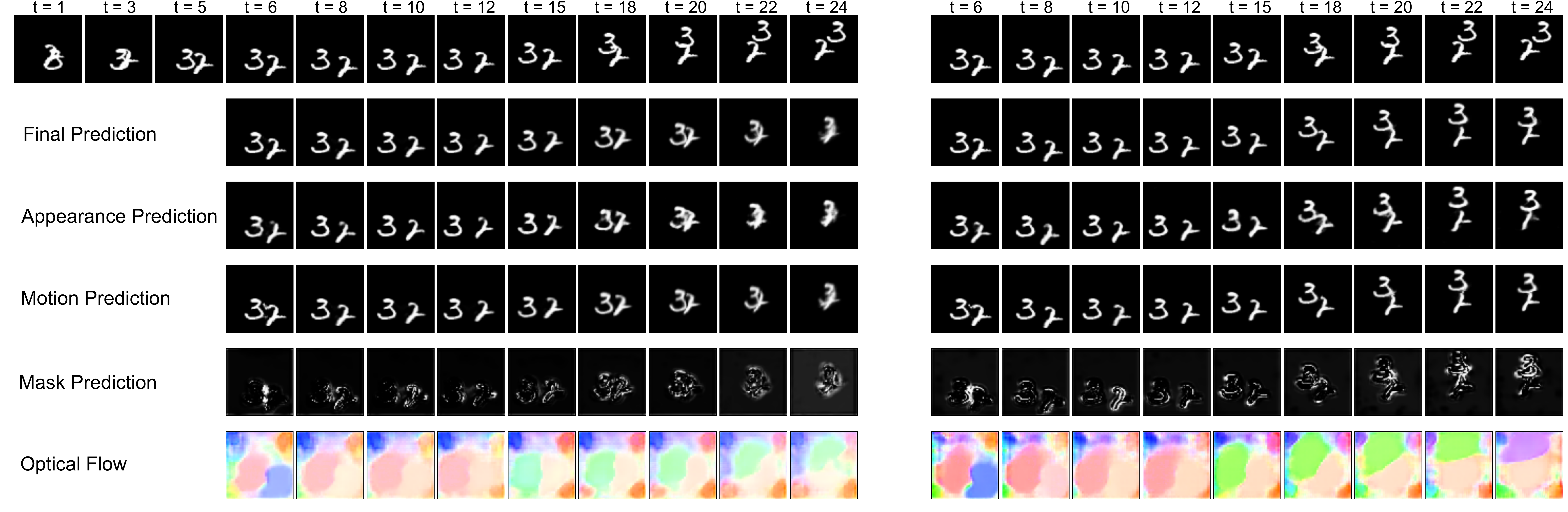}
    \caption{\textbf{SLAMP-Baseline (left) \vs SLAMP (right) on MNIST.} The top row shows the ground truth, followed by the frame predictions by the final, the appearance, the motion, and the last two rows show the mask and the optical flow predictions with false coloring. In this challenging case with bouncing and collisions, the baseline confuses the digits and cannot predict last frames correctly whereas SLAMP can generate predictions very close to the ground truth by learning smooth transitions in the motion history, as can be seen from optical flow predictions. See \figref{fig:color_wheel} for the color wheel showing the direction of flow.}
    \label{fig:qual_mnist_comp}
    \vspace{-4mm}
\end{figure*}
We evaluate the performance of the proposed approach and compare it to the previous methods on three standard video prediction datasets including Stochastic Moving MNIST, KTH Actions \cite{Schuldt2004CVPR} and BAIR Robot Hand \cite{EbertFLL17}. 
We specifically compare our baseline model (SLAMP-Baseline) and our model (SLAMP) to SVG \cite{Denton2018ICML} which is a special case of our baseline with a single pixel decoder, SAVP \cite{Lee2018ARXIV}, SV2P \cite{Babaeizadeh2018ICLR}, and lastly to SRVP \cite{Franceschi2020ICML}. 
We also compare our model to SVG \cite{Denton2018ICML} and SRVP \cite{Franceschi2020ICML} on two different challenging real world datasets, KITTI \cite{Geiger2012CVPR, Geiger2013IJRR} and Cityscapes \cite{Cordts2016CVPR}, with moving background and complex object motion.
We follow the evaluation setting introduced in \cite{Denton2018ICML} 
by generating 100 samples for each test sequence and report the results according to the best one in terms of average performance over the frames.
Our experimental setup including training details and parameter settings can be found in Appendix \secref{sec:training_details}. We also share the code for reproducibility.

\begin{figure}[ht]
    \centering
    \includegraphics[width=\linewidth]{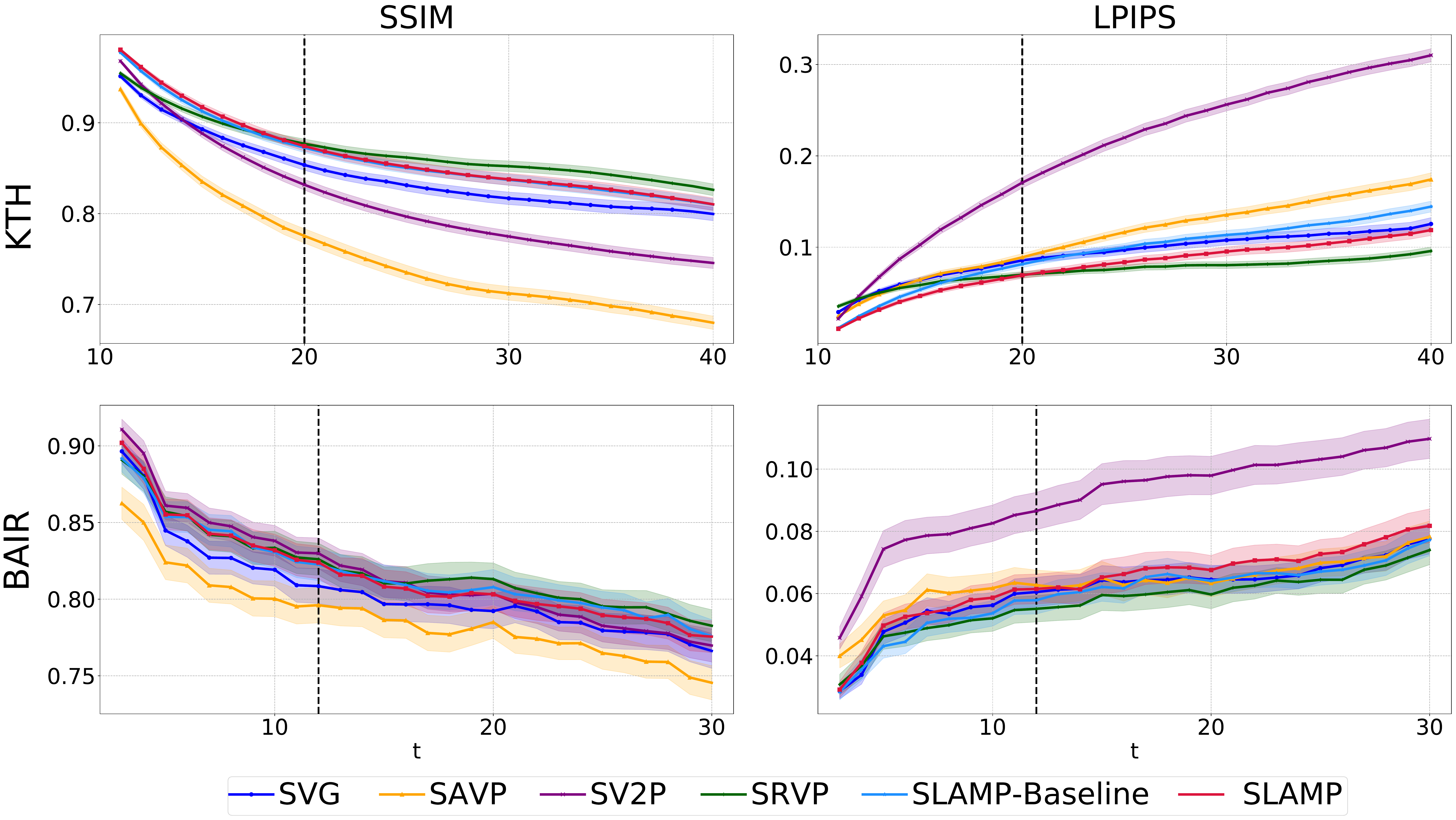}
    \caption{\textbf{Quantitative Results on KTH and BAIR.} We compare our results to previous work in terms of PSNR, SSIM, and LPIPS metrics with respect to the time steps on KTH (\textbf{top}), and BAIR (\textbf{bottom}) datasets, with 95\%-confidence intervals. Vertical bars mark the length of training sequences. SLAMP outperforms previous work including SVG \cite{Denton2018ICML}, SAVP \cite{Lee2018ARXIV}, SV2P \cite{Babaeizadeh2018ICLR} and performs comparably to the state of the art method SRVP \cite{Franceschi2020ICML} on both datasets.}
    \label{fig:results_kth_bair}
    \vspace{-3mm}
\end{figure}

\begin{table}[h!]
    \captionsetup[table]{justification=centering}
    \sisetup{detect-weight, table-align-uncertainty=true, mode=text}
    \renewrobustcmd{\bfseries}{\fontseries{b}\selectfont}
    \renewrobustcmd{\boldmath}{}
    \caption{
        \label{tab:fvd}
        \textbf{FVD Scores on KTH and BAIR.} This table compares all the methods in terms of FVD scores with their $95\%$-confidence intervals over five different samples from the models. Our model is the second best on KTH and among top three methods on BAIR.}
    \centering
    \vspace{-0.05in}
    \small
    \begin{tabular}{lcc}
        \toprule
        Dataset & KTH & BAIR \tabularnewline
        \midrule
        SV2P & 636 $\pm$ 1 & 965 $\pm$ 17 \tabularnewline
        SAVP & 374 $\pm$ 3 & \bfseries 152 $\pm$ 9 \tabularnewline
        SVG & 377 $\pm$ 6 & 255 $\pm$ 4 \tabularnewline
        SRVP & \bfseries 222 $\pm$ 3 & \underline{163}$\pm$ 4 \tabularnewline
        SLAMP-Baseline & 236 $\pm$ 2 & 245 $\pm$ 5 \tabularnewline
        SLAMP & \underline{228} $\pm$ 5 & {\textemdash} \tabularnewline
        \bottomrule
    \end{tabular}
    \vspace{-2mm}
\end{table}

\boldparagraph{Evaluation Metrics} We compare the performance using three frame-wise metrics and a video-level one.
Peak Signal-to-Noise Ratio (PSNR), \emph{higher better}, based on $L_2$ distance between the frames penalizes differences in dynamics but also favors blur predictions.
Structured Similarity (SSIM), \emph{higher better}, compares local patches to measure similarity in structure spatially. 
Learned Perceptual Image Patch Similarity (LPIPS) \cite{Zhang2018CVPR}, \emph{lower better}, measures the distance between learned features extracted by a CNN trained for image classification.
Frechet Video Distance (FVD) \cite{Unterthiner2019ARXIV}, \emph{lower better}, compares temporal dynamics of generated videos to the ground truth in terms of representations computed for action recognition.

\boldparagraph{Stochastic Moving MNIST} This dataset contains up to two MNIST digits moving linearly and bouncing from walls with a random velocity as introduced in \cite{Denton2018ICML}.
Following the same training and evaluation settings as in the previous work, we condition on the first 5 frames during training and learn to predict the next 10 frames. During testing, we again condition on the first 5 frames but predict the next 20 frames.

\figref{fig:results_mnist} shows quantitative results on MNIST in comparison to SVG \cite{Denton2018ICML} and SRVP \cite{Franceschi2020ICML} in terms of PSNR and SSIM, omitting LPIPS as in SRVP. Our baseline model with a motion decoder (SLAMP-Baseline) already outperforms SVG on both metrics. SLAMP further improves the results by utilizing the motion history and reaches a comparable performance to the state of the art model SRVP. This shows the benefit of separating the video into static and dynamic parts in both state-space models (SRVP) and auto-regressive models (ours, SLAMP). This way, models can better handle challenging cases such as crossing digits as shown next.

We qualitatively compare SLAMP to SLAMP-Baseline on MNIST in \figref{fig:qual_mnist_comp}. The figure shows predictions of static and dynamic parts as appearance and motion predictions, as well the final prediction as the combination of the two. According to the mask prediction, the final prediction mostly relies on the dynamic part shown as black on the mask and uses the static component only near the motion boundaries. Moreover, optical flow prediction does not fit the shape of the digits but expands as a region until touching the motion region of the other digit.
This is due to the uniform black background. Moving a black pixel in the background randomly is very likely to result in another black pixel in the background, which means zero-loss for the warping result.
Both models can predict optical flow correctly for the most part and resort to the appearance result in the occluded regions. However, continuity in motion is better captured by SLAMP with the colliding digits whereas the baseline model cannot recover from it, leading to blur results, far from the ground truth. Note that we pick the best sample for both models among 100 samples according to LPIPS.

\boldparagraph{KTH Action Dataset}
\begin{figure}[!t]
    \centering
    \includegraphics[width=\linewidth]{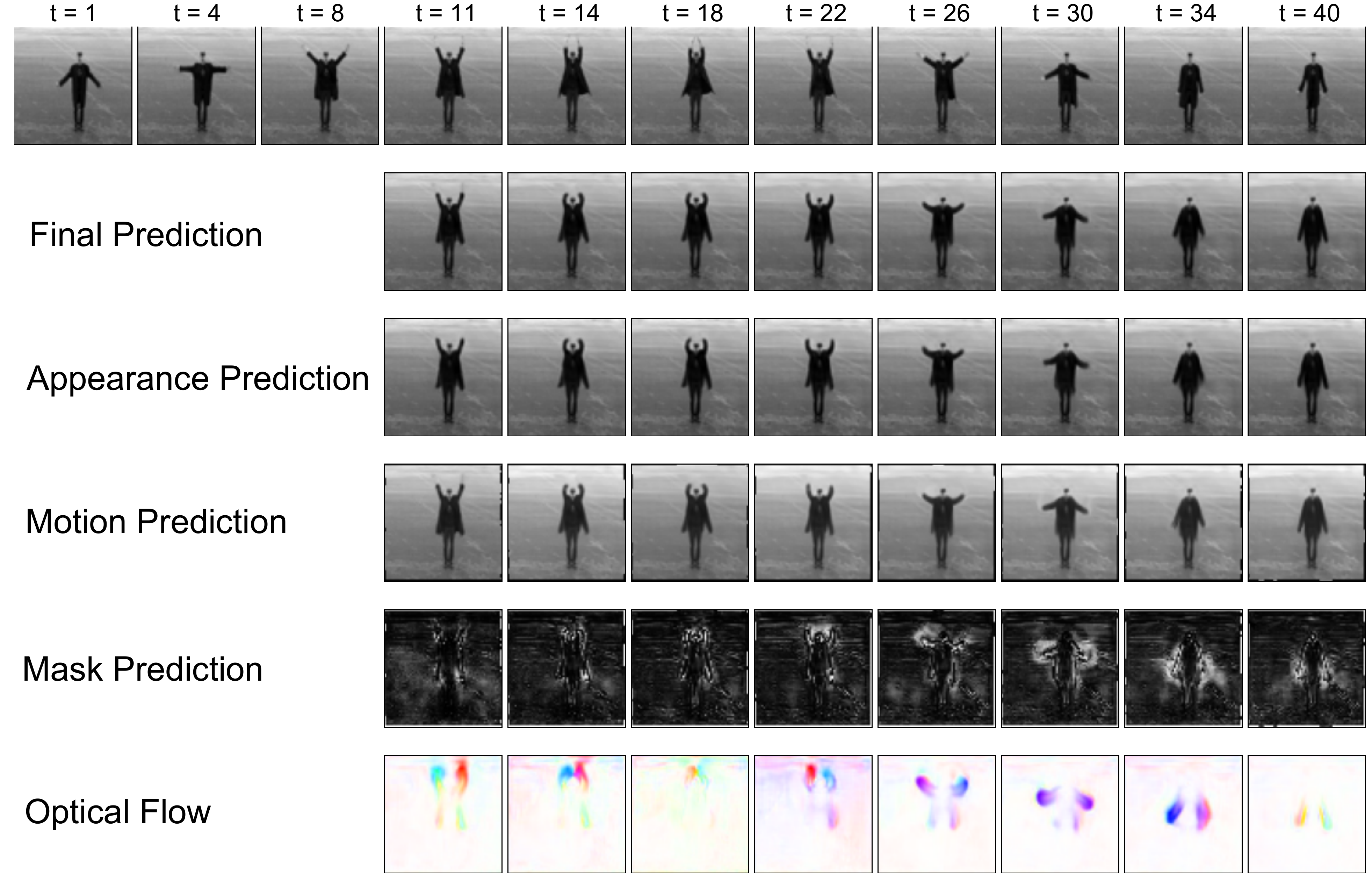}
    \caption{\textbf{Qualitative Results on KTH} We visualize the results of SLAMP on KTH dataset. The top row shows the ground truth, followed by the frame predictions by the final, the appearance, the motion, and the last two rows show the mask and the optical flow predictions. The mask prediction combines the appearance prediction (white) and the motion prediction (black) into the final prediction.}
    \label{fig:qual_kth_hist_bair_motion}
    \vspace{-5mm}
\end{figure}
\begin{figure*}[ht]
    \centering
    \includegraphics[width=\textwidth]{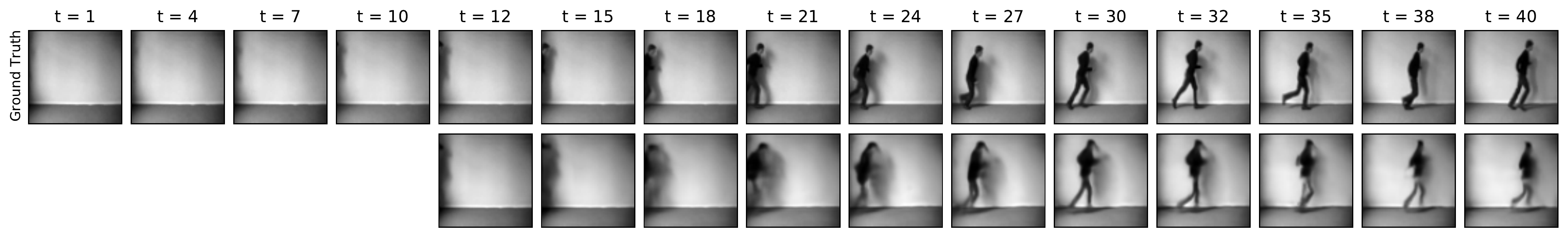}
    \caption{\textbf{Subject Appearing after the Conditioning Frames.} This figure shows a case where the subject appears after conditioning frames on KTH with ground truth (\textbf{top}) and a generated sample by our model (\textbf{bottom}). This shows our model's ability to capture dynamics of the dataset by generating samples close to the ground truth, even conditioned on empty frames.}
    \label{fig:qual_kth_hist_no_person}
    \vspace{-2mm}
\end{figure*}
KTH dataset contains real videos where people perform a single action such as walking, running, boxing, \etc in front of a static camera~\cite{Schuldt2004CVPR}. We expect our model with motion history to perform very well by exploiting regularity in human actions on KTH.
Following the same training and evaluation settings used in the previous work, we condition on the first 10 frames and learn to predict the next 10 frames. During testing, we again condition on the first 10 frames but predict the next 30 frames. 

\figref{fig:results_kth_bair} and \tabref{tab:fvd} show quantitative results on KTH in comparison to previous approaches. Both our baseline and SLAMP models outperform previous approaches and perform comparably to SRVP, in all metrics including FVD. A detailed visualization of all three frame predictions as well as flow and mask are shown in \figref{fig:qual_kth_hist_bair_motion}. Flow predictions are much more fine-grained than MNIST by capturing fast motion of small objects such as hands or thin objects such as legs (see Appendix~\secref{sec:add_qual}). The mask decoder learns to identify regions around the motion boundaries which cannot be matched with flow due to occlusions and assigns more weight to the appearance prediction in these regions.

On KTH, the subject might appear after the conditioning frames. These challenging cases can be problematic for some previous work as shown in SRVP~\cite{Franceschi2020ICML}. Our model can generate samples close to the ground truth despite very little information on the conditioning frames as shown in \figref{fig:qual_kth_hist_no_person}. The figure shows the best sample in terms of LPIPS, please see Appendix~\secref{sec:add_qual} for a diverse set of samples with subjects of various poses appearing at different time steps.

\boldparagraph{BAIR Robot Hand}
This dataset contains videos of a robot hand moving and pushing objects on a table \cite{EbertFLL17}. Due to uncertainty in the movements of the robot arm, 
BAIR is a standard dataset for evaluating stochastic video prediction models.
Following the training and evaluation settings used in the previous work, we condition on the first 2 frames and learn to predict the next 10 frames. During testing, we again condition on the first 2 frames but predict the next 28 frames.

\begin{figure*}[ht!]
\centering
\begin{subfigure}[b]{0.99\textwidth}
   \includegraphics[width=1\linewidth]{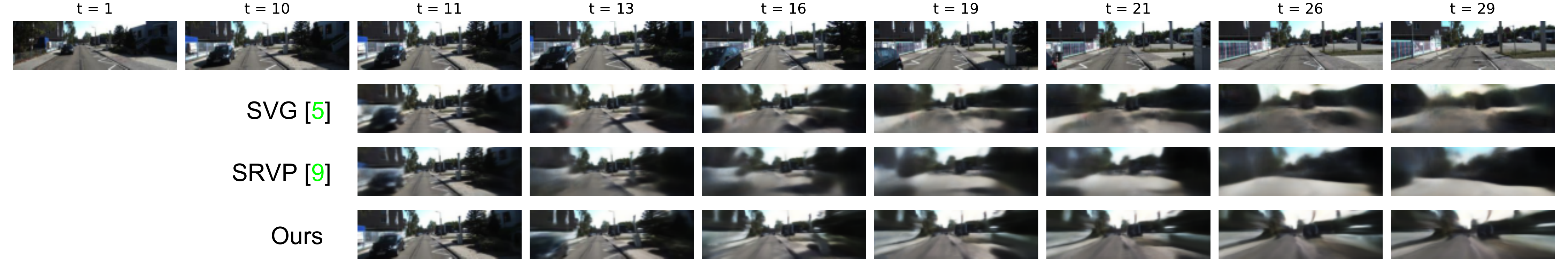}
   \vspace{-4.5mm}
\end{subfigure}
\begin{subfigure}[b]{0.99\textwidth}
   \includegraphics[width=1\linewidth]{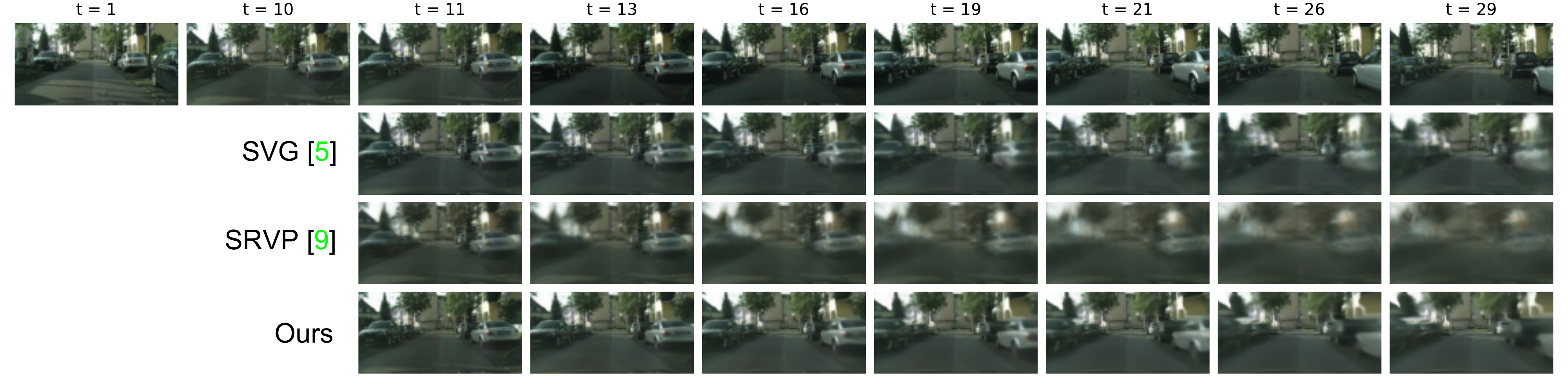}
\end{subfigure}
\vspace{-2.5mm}
\caption{\textbf{Qualitative Comparison.} We compare SLAMP to SVG~\cite{Denton2018ICML} and SRVP~\cite{Franceschi2020ICML} on KITTI~(\textbf{top}) and Cityscapes~(\textbf{bottom}). Our model can better capture the changes due to ego-motion thanks to explicit modeling of motion history.}
\label{fig:kitti_city_qual}
\vspace{-4.5mm}
\end{figure*}

We show quantitative results on BAIR in \figref{fig:results_kth_bair} and \tabref{tab:fvd}. Our baseline model achieves comparable results to SRVP, outperforming other methods in all metrics except SV2P~\cite{Babaeizadeh2018ICLR} in PSNR and SAVP~\cite{Lee2018ARXIV} in FVD. With 2 conditioning frames only, SLAMP cannot utilize the motion history and performs similarly to the baseline model on BAIR (see Appendix~\secref{sec:quant_res}). This is simply due to the fact that there is only one flow field to condition on, in other words, no motion history. Therefore, we only show the results of the baseline model on this dataset.

\begin{table}[t!]
    \sisetup{detect-weight, table-align-uncertainty=true, mode=text}
    \renewrobustcmd{\bfseries}{\fontseries{b}\selectfont}
    \renewrobustcmd{\boldmath}{}
    \centering
    \label{tab:kitti_city_results}
    \caption{\textbf{Results with a Moving Background.} We evaluate our model SLAMP in comparison to SVG and SRVP on 
    KITTI~\cite{Geiger2013IJRR} and Cityscapes~\cite{Cordts2016CVPR} datasets by conditioning on 10 frames and predicting 20 frames into the future.}
    \label{tab:kitti_city_results}
    \begin{tabular}{lccc}
        \toprule
        Models &  PSNR ($\uparrow$) & SSIM ($\uparrow$) & LPIPS ($\downarrow$) \tabularnewline 
        \midrule
        SVG~\cite{Denton2018ICML} & 12.70 $\pm$ 0.70 & 0.329 $\pm$ 0.030 &  \underline{0.594} $\pm$ 0.034 
        \tabularnewline
        SRVP~\cite{Franceschi2020ICML}  & \underline{13.41} $\pm$ 0.42 & \underline{0.336} $\pm$ 0.034 & 0.635 $\pm$ 0.021 \tabularnewline
        SLAMP & \bfseries 13.46 $\pm$ 0.74 & \bfseries 0.337 $\pm$ 0.034 & \bfseries 0.537 $\pm$ 0.042
        \tabularnewline       \bottomrule
    \end{tabular}
    \subcaption*{KITTI~\cite{Geiger2012CVPR, Geiger2013IJRR} 
    }
    \begin{tabular}{lccc}
        \toprule
        Models &  PSNR ($\uparrow$) & SSIM ($\uparrow$) & LPIPS ($\downarrow$) \tabularnewline 
        \midrule
        SVG~\cite{Denton2018ICML} & 20.42 $\pm$ 0.63 & \underline{0.606} $\pm$ 0.023 & \underline{0.340} $\pm$ 0.022
        \tabularnewline
        SRVP~\cite{Franceschi2020ICML}  & \underline{20.97} $\pm$ 0.43 & 0.603 $\pm$ 0.016 & 0.447 $\pm$ 0.014 \tabularnewline
        SLAMP & \bfseries 21.73 $\pm$ 0.76 & \bfseries 0.649 $\pm$ 0.025 & \bfseries 0.2941 $\pm$ 0.022
        \tabularnewline       \bottomrule
    \end{tabular}
    \subcaption*{Cityscapes~\cite{Cordts2016CVPR}}
    \vspace{-5mm}
\end{table}

\boldparagraph{Real-World Driving Datasets} We perform experiments on two challenging autonomous driving datasets: KITTI~\cite{Geiger2012CVPR, Geiger2013IJRR} and Cityscapes~\cite{Cordts2016CVPR} with various challenges.
Both datasets contain everyday real-world scenes with complex dynamics due to both background and foreground motion. KITTI is recorded in one town in Germany while Cityscapes is recorded in 50 European cities, leading to higher diversity.

Cityscapes primarily focuses on semantic understanding of urban street scenes, therefore contains a larger number of dynamic foreground objects compared to KITTI. However, motion lengths are larger on KITTI due to lower frame-rate. On both datasets, we condition on 10 frames and predict 10 frames into the future to train our models. Then at test time, we predict 20 frames conditioned on 10 frames.

As shown in \tabref{tab:kitti_city_results}, SLAMP outperforms both methods on all of the metrics on both datasets, which shows its ability to generalize to the sequences with moving background. Even SVG \cite{Denton2018ICML} performs better than the state of the art SRVP~\cite{Franceschi2020ICML} in LPIPS metric for KITTI and on both SSIM and LPIPS for Cityscapes, which shows the limitations of SRVP on scenes with dynamic backgrounds. We also perform a qualitative comparison to these methods in \figref{fig:teaser} and \figref{fig:kitti_city_qual}. SLAMP can better preserve the scene structure thanks to explicit modeling of ego-motion history in the background.

\boldparagraph{Visualization of Latent Space}
We visualize stochastic latent variables of the dynamic component on KTH compared to the static and SVG. (see \figref{fig:tsne-both} and \figref{fig:tsne-svg})
\vspace{-2mm}

\section{Conclusion}
\label{sec:conc}
We presented a stochastic video prediction framework to decompose video content into appearance and dynamic components. 
Our baseline model with deterministic motion and mask decoders outperforms SVG, which is a special case of our baseline model. 
Our model with motion history, SLAMP, further improves the results and reaches the performance of the state of the art method SRVP on the previously used datasets. Moreover, it outperforms both SVG and SRVP on two real-world autonomous driving datasets with dynamic background and complex motion.
We show that motion history enriches model's capacity to 
predict future, leading to better predictions in challenging cases. 

Our model with motion history cannot realize its full potential in standard settings of stochastic video prediction datasets. A fair comparison is not possible on BAIR due to the little number of conditioning frames. BAIR holds a great promise with changing background but infrequent, small changes are not reflected in current evaluation metrics.

An interesting direction is stochastic motion decomposition, maybe with hierarchical latent variables, for modelling camera motion and motion of each object in the scene separately.

\vspace{-2mm}
\paragraph{Acknowledgements.} 
{
We would like to thank Jean-Yves Franceschi and Edouard Delasalles for providing technical and numerical details for the baseline performances, and Deniz Yuret for helpful discussions and comments. 
K. Akan was supported by KUIS AI Center fellowship, F. G\"uney by TUBITAK 2232 International Fellowship for Outstanding Researchers Programme, E. Erdem in part by GEBIP 2018 Award of the Turkish Academy of Sciences, A. Erdem by  BAGEP 2021 Award of the Science Academy.
}

\clearpage
\bibliography{bibliography_long, ref}
\bibliographystyle{iccv}

\onecolumn
\begin{appendix}
In this part, we provide additional illustrations, derivations, and results for our paper ``SLAMP: Stochastic Latent Appearance and Motion Prediction". 
We first show the model illustrations of our proposed model (SLAMP) and our baseline model (SLAMP-Baseline) in comparison to the previous work by Denton et al. \cite{Denton2018ICML} (SVG) in \secref{sec:model_figs}. 
In \secref{sec:deriv}, we provide the full derivations of the variational inference, evidence lower bounds of our baseline model (\secref{sec:deriv_slamp_baseline}) and our proposed model (\secref{sec:deriv_slamp}). 
We explain the architectural choices and training details in \secref{sec:training_details}. 
In \secref{sec:quant_res}, we present detailed versions of the quantitative results in the main paper. In addition, we present the ablation experiments for our model's mask component. We evaluate and compare the predictions of static, dynamic heads of the model, simple averaging of the two without a mask, and our full model with learned mask.
In \secref{sec:add_qual}, we first provide the color wheel to interpret optical flow predictions, and a comparison of static and dynamic latent variables. 
We then present several qualitative results both with details as in the main paper and with random samples, showing the diversity of the generated samples on all datasets.

For video examples, please visit \url{https://kuis-ai.github.io/slamp/}.
\section{Model Illustrations}
\label{sec:model_figs}
\begin{figure}[h]
\centering
\includegraphics[width=.7\textwidth]{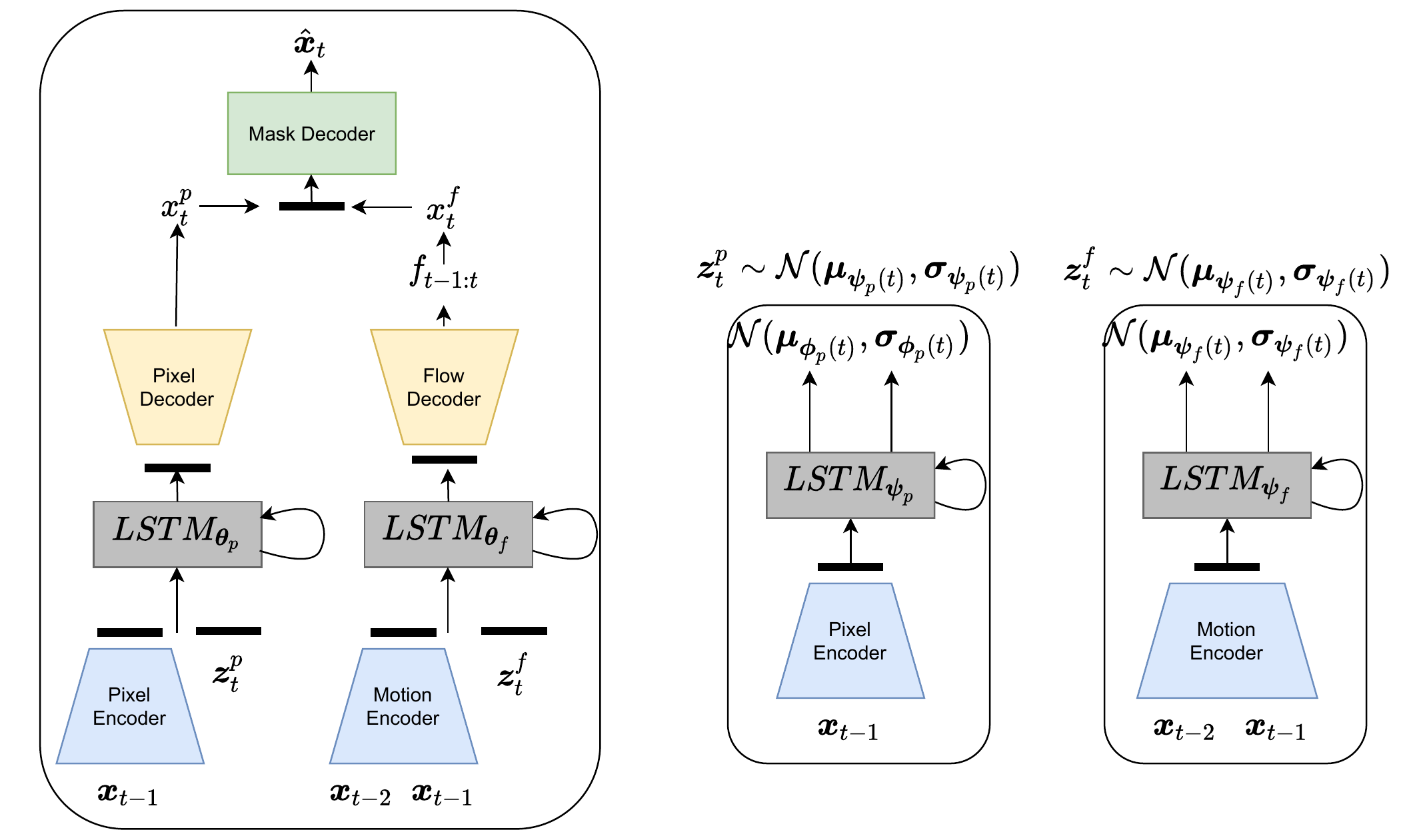}
\caption{\textbf{Illustration of the Inference Procedure for SLAMP.} This figure illustrates the difference between the inference time and the train time in terms of the distributions the latent variables are sampled from. While at train time, latent variables are sampled from the posterior distribution, at test time, they are sampled from the posterior for the conditioning frames and from the prior for the following frames.
}
\label{fig:illustration_slamp_inf}
\end{figure}

\begin{figure}[h]
\centering
\includegraphics[width=0.48\textwidth]{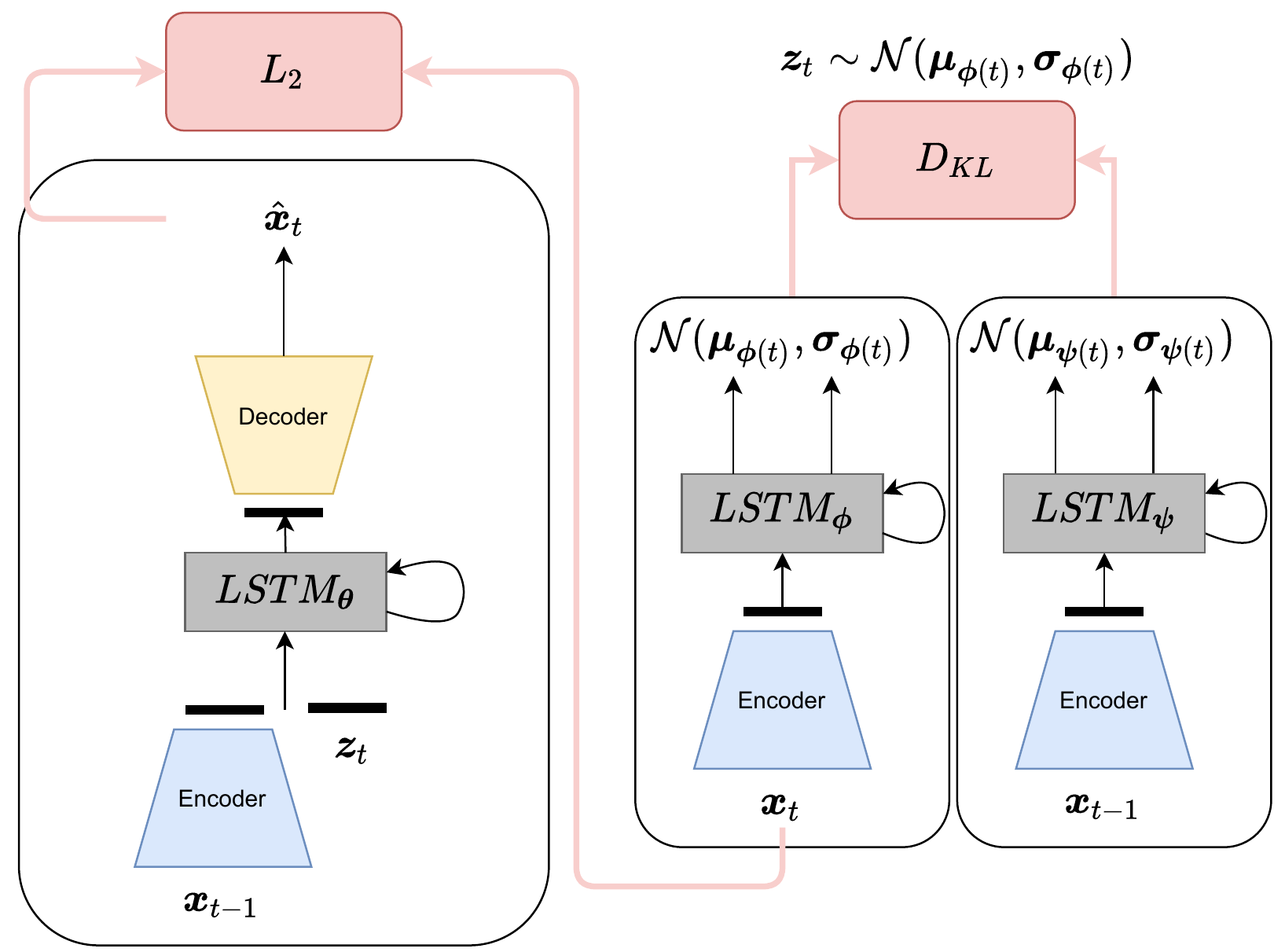} $\quad$
\includegraphics[width=0.48\textwidth]{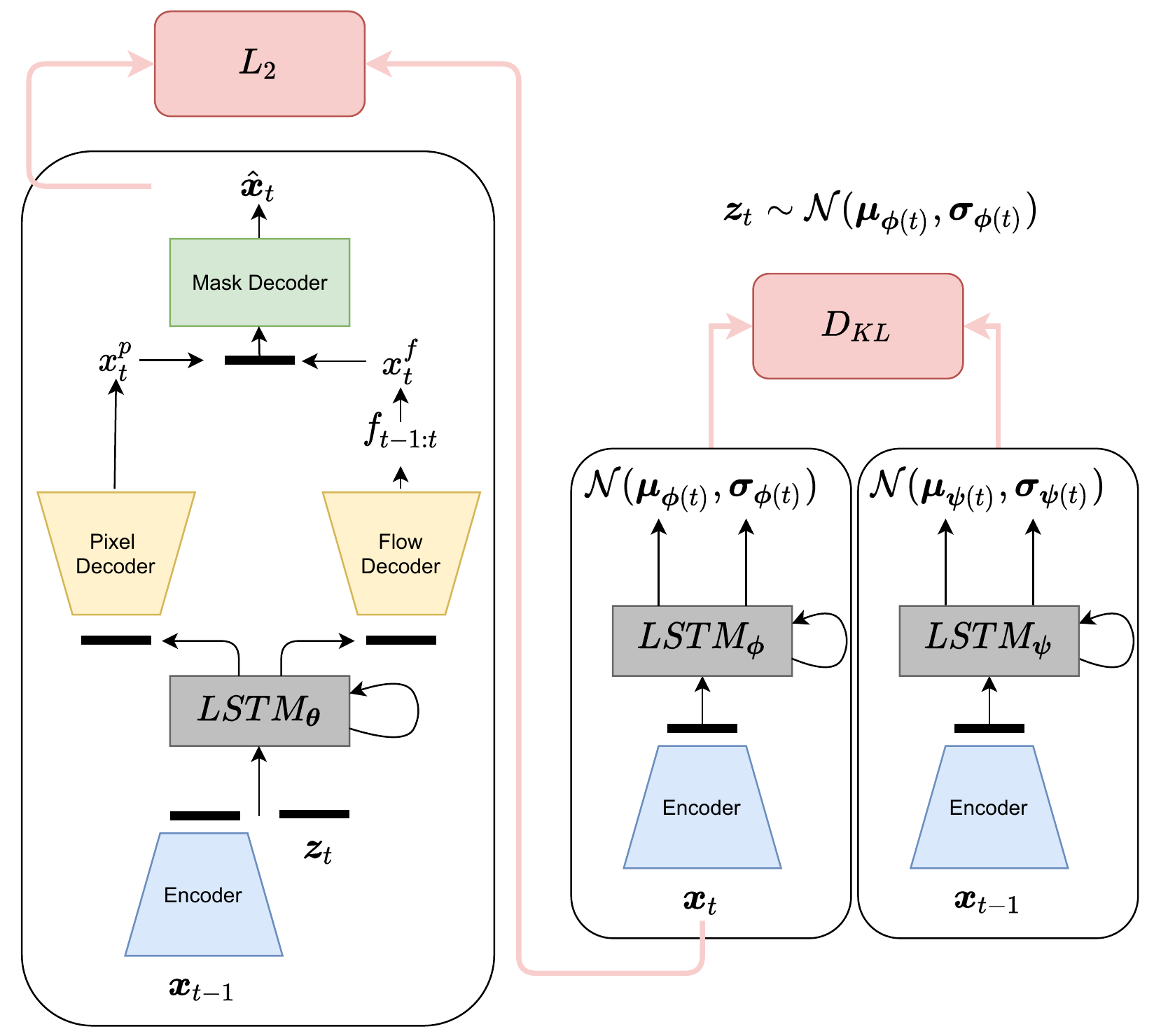}
\caption{\textbf{Illustration of the Training Procedure for SVG (left) and SLAMP-Baseline (right).} The main difference between SVG and SLAMP-Baseline is that SLAMP-Baseline has three decoders instead of one pixel decoder. In SLAMP-Baseline, in addition to the appearance prediction $\bx_t^p$, we also estimate flow $\bff_{t-1:t}$ and warp the previous frame according to the estimated flow to obtain motion prediction $\bx_t^f$. Mask decoder takes appearance and motion predictions as input and generates a weighted combination of the two, $\hat{\bx}_t$ as the final prediction. Note that SVG corresponds to only appearance prediction case of our baseline model.}
\label{fig:illustration_others_train}
\end{figure}

\begin{figure}[h]
\centering
\includegraphics[width=0.48\textwidth]{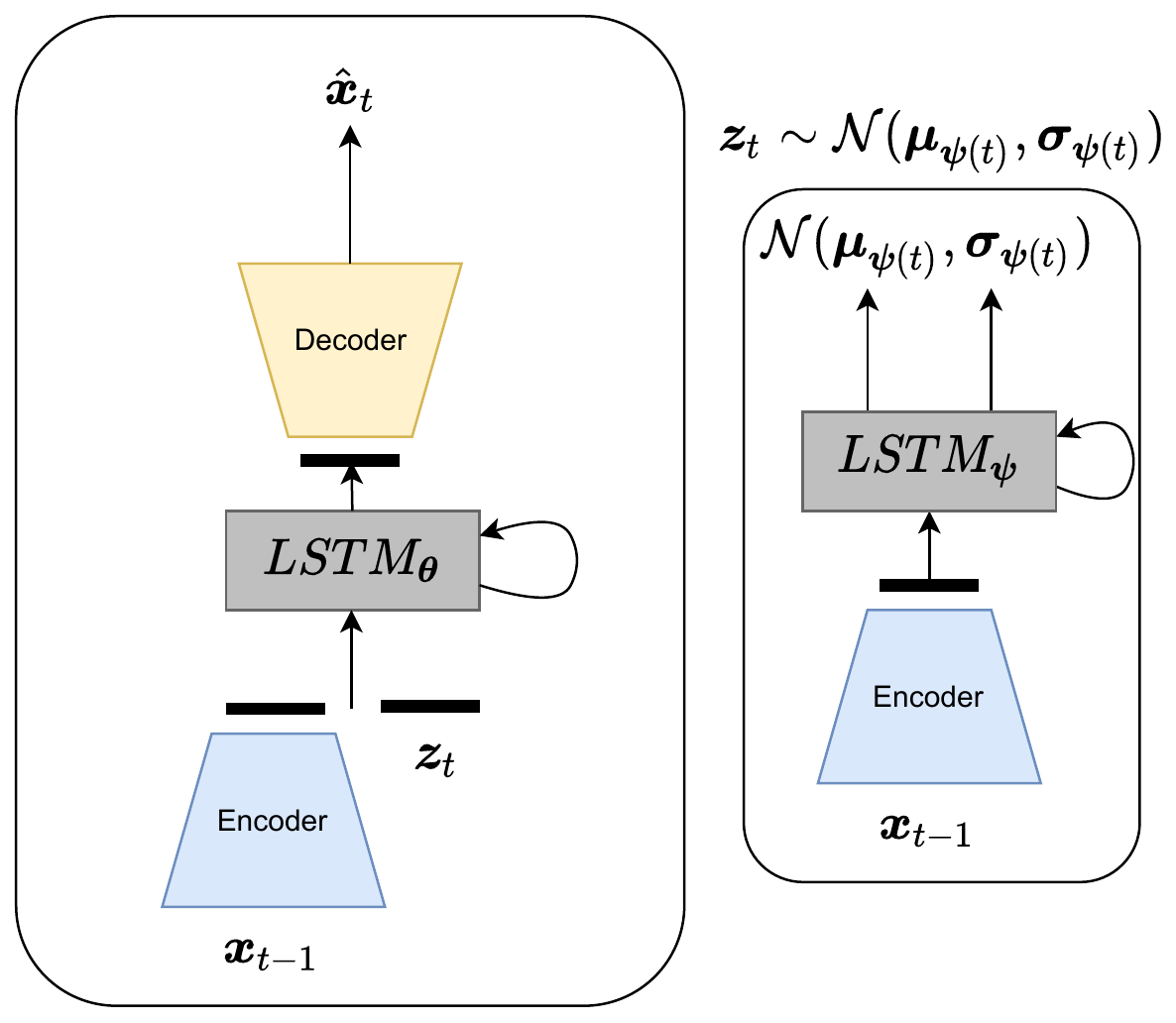} $\quad$
\includegraphics[width=0.48\textwidth]{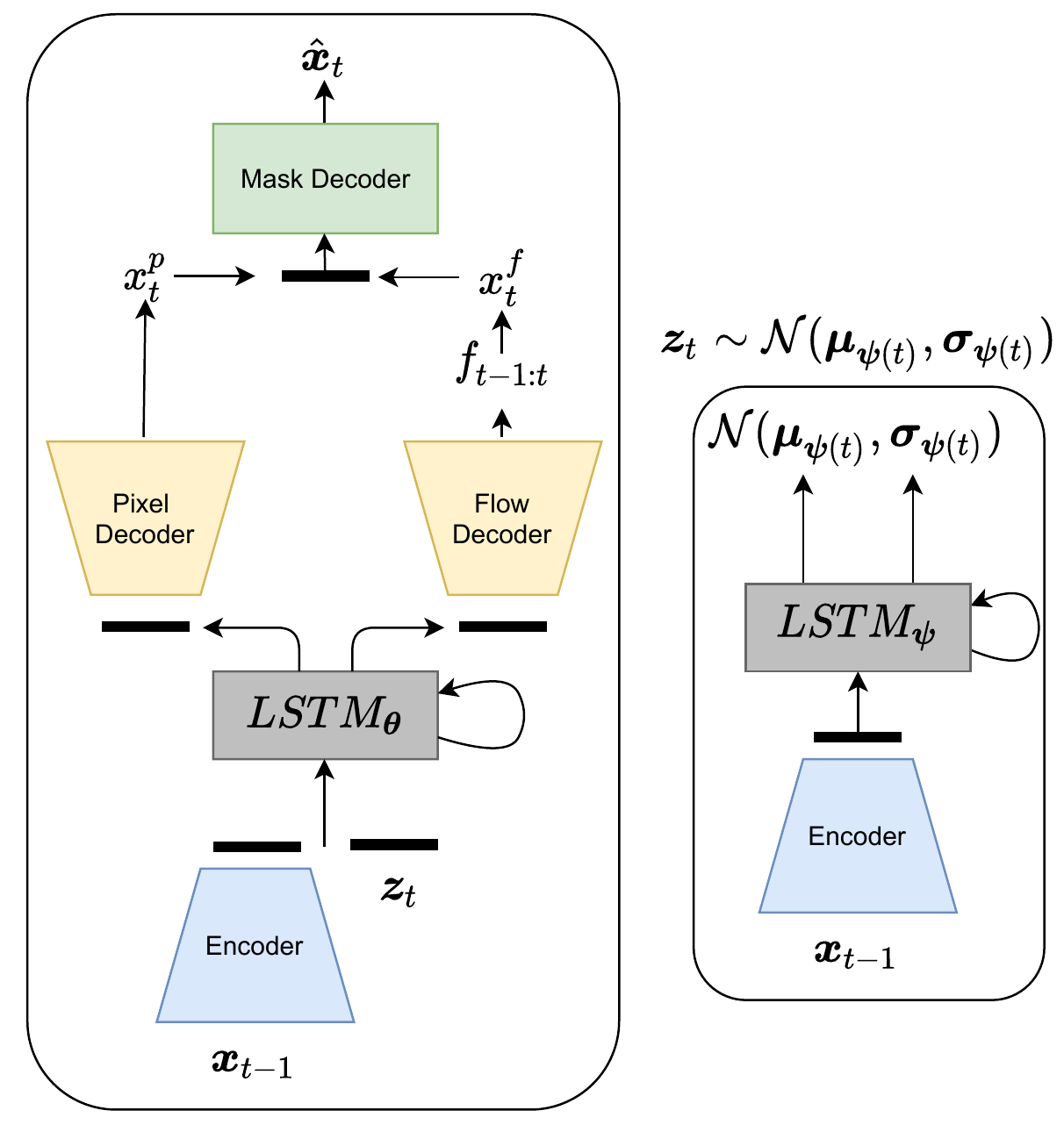}
\caption{\textbf{Illustration of the Inference Procedure for SVG (left) and SLAMP-Baseline (right).} This figure illustrates the inference time in comparison to the train time in terms of the distribution the latent variables are sampled from. While at train time, latent variables are sampled from the posterior distribution, at test time, they are sampled from the posterior for the conditioning frames and from the prior for the following frames.}
\label{fig:illustration_others_inf}
\end{figure}

In \figref{fig:illustration_slamp_inf}, we provide the inference procedure of our model SLAMP, in addition to the training procedure provided in the main paper.
Moreover, we present graphical illustrations of the training (\figref{fig:illustration_others_train}) and the inference procedures (\figref{fig:illustration_others_inf}) of our baseline model, SLAMP-Baseline, in comparison to SVG \cite{Denton2018ICML}. 

\clearpage
\section{Derivations}
\label{sec:deriv}

Here, we provide derivations of inference steps and variational lower bounds of the baseline method, SLAMP-Baseline (\secref{sec:deriv_slamp_baseline}), and our method SLAMP (\secref{sec:deriv_slamp}).
\subsection{Derivation of the ELBO for SLAMP-Baseline}
\label{sec:deriv_slamp_baseline}
We first derive the variational lower bound for the baseline model with one posterior and one learned prior distribution. 
\begin{eqnarray}
\log p_{\btheta}(\bx) &=& \log \int\limits_{\bz} p_{\btheta}(\bx \vert \bz)~p(\bz \vert \bx) \\ 
&=& \log \int\limits_{\bz} p_{\btheta}(\bx \vert \bz)~p(\bz \vert \bx)~\frac{q_{\bphi}(\bz \vert \bx)}{q_{\bphi}(\bz \vert \bx)} \nonumber \\
&=& \log \mathbb{E}_{q_{\bphi}(\bz \vert \bx)} \frac{p_{\btheta}(\bx \vert \bz)~p(\bz \vert \bx)}{q_{\bphi}(\bz \vert \bx)} \nonumber \\
&\ge& \mathbb{E}_{q_{\bphi}(\bz \vert \bx)} \log \frac{p_{\btheta}(\bx \vert \bz)~p(\bz \vert \bx) }{q_{\bphi}(\bz \vert \bx)} \nonumber \\
&=& \mathbb{E}_{q_{\bphi}(\bz \vert \bx)} \log p_{\btheta}(\bx \vert \bz) - \mathbb{E}_{q_{\bphi}(\bz \vert \bx)} \log \frac{q_{\bphi}(\bz \vert \bx)}{p(\bz \vert \bx)} \nonumber \\
&=& \mathbb{E}_{q_{\bphi}(\bz \vert \bx)} \log p_{\btheta}(\bx \vert \bz) - \KLDD{q_{\bphi}(\bz \vert \bx)}{p(\bz \vert \bx)} \nonumber
\end{eqnarray}

We model the posterior distribution with a recurrent network. The recurrent network outputs a different posterior distribution, $q_{\bphi}(\bz_t \vert \bx_{1:t})$, at every time step. Due to independence of latent variables across time, $\bz = [\bz_1, \bz_2, \dots, \bz_T]$, we can derive the estimation of posterior distribution across time steps as follows:
\begin{align}
    q_{\bphi}(\bz \vert \bx) = \prod\limits_t q_{\bphi}(\bz_t \vert \bx_{1:t})
\end{align}

Since the latent variables, $\bz = [\bz_1, \bz_2, \cdots, \bz_T]$, are independent across time, we can further decompose Kullback-Leibler term in the evidence lower bound into individual time steps:
\begin{align}
\label{eq:slamp_baseline_kl_timesptes}
&\KLD{q_{\bphi}(\bz \vert \bx)}{p(\bz \vert \bx_{1:t-1})} \\
&= \int\limits_\bz q_{\bphi}(\bz \vert \bx)~\log \frac{q_{\bphi}(\bz \vert \bx)}{p(\bz \vert \bx_{1:t-1})} \nonumber \\
&=\int\limits_{\bz_1} \cdots \int\limits_{\bz_T} q_{\bphi}(\bz_1 \vert \bx_1) \cdots q_{\bphi}(\bz_T \vert \bx_{1:T})~\log \frac{q_{\bphi}(\bz_1 \vert \bx_1) \cdots q_{\bphi}(\bz_t \vert \bx_{1:T})}{p(\bz_1 \vert \bx_{1}) \cdots p(\bz_T \vert \bx_{1:T-1})} \nonumber \\
&=\int\limits_{\bz_1} \cdots \int\limits_{\bz_T} q_{\bphi}(\bz_1 \vert \bx_1) \cdots q_{\bphi}(\bz_T \vert \bx_{1:T})~\sum\limits_t \log \frac{q_{\bphi}(\bz_t \vert \bx_{1:t})}{p(\bz_t \vert \bx_{1:t-1})} \nonumber \\
&=\sum\limits_t \int\limits_{\bz_1} \cdots \int\limits_{\bz_T} q_{\bphi}(\bz_1 \vert \bx_1) \cdots q_{\bphi}(\bz_T \vert \bx_{1:T})~\log \frac{q_{\bphi}(\bz_t \vert \bx_{1:t})}{p(\bz_t \vert \bx_{1:t-1})} \nonumber \\
&\text{And because } \int\limits_x p(x) = 1 \text{, this simplifies to:} \nonumber \\
&= \sum\limits_t \int\limits_{\bz_t} q_{\bphi}(\bz_t \vert \bx_{1:t})~\log \frac{q_{\bphi}(\bz_t \vert \bx_{1:t})}{p(\bz_t \vert \bx_{1:t-1})} \nonumber \nonumber \\
&= \sum\limits_t \KLD{q_{\bphi}(\bz_t \vert \bx_{1:t})}{p(\bz_t \vert \bx_{1:t-1})} \nonumber
\end{align}

At each time step, our model predicts $\bx_{t}$, conditioned on $\bx_{t-1}$ and $\bz_{t}$. Since our model has recurrence connections, it considers not only $\bx_{t-1}$ and $\bz_{t}$, but also $\bx_{1:t-2}$ and $\bz_{1:t-1}$. Therefore, we can further write our inference as:
\begin{align}
    \log p_{\btheta}(\bx \vert \bz) &= \log \prod\limits_t p_{\btheta}(\bx_t \vert \bx_{1:t-1}, \bz_{1:t})\\
    &= \sum\limits_t \log p_{\btheta}(\bx_t \vert \bx_{1:t-1}, \bz_{1:t}) \nonumber
\end{align}

Combining all of them leads to the following variational lower bound:
\begin{align}
\log p_{\btheta}(\bx) &\geq \mathcal{L}_{\btheta, \bphi, \bpsi}(\bx_{1:T}) \\
&=  \mathbb{E}_{q_{\bphi}(\bz \vert \bx)} \log p_{\btheta}(\bx \vert \bz) -  \KLDD{q_{\bphi}(\bz \vert \bx)}{p_{\bpsi}(\bz \vert \bx)} \nonumber \\
&=  \sum\limits_t \big[ \mathbb{E}_{q_{\bphi}(\bz_{1:t} \vert \bx_{1:t})}   \log p_{\btheta}(\bx_t \vert \bx_{1:t-1}, \bz_{1:t}) \nonumber \\
& \hspace{13mm} - \KLDD{q_{\bphi}(\bz_t \vert \bx_{1:t}}{p_{\bpsi}(\bz_t \vert \bx_{1:t-1})} \big] \nonumber
\end{align}

\subsection{Derivation of the ELBO for SLAMP}
\label{sec:deriv_slamp}
In this section, we derive the variational lower bound for the proposed model with two posterior and two learned prior distributions. 
\begin{align}
\log p_{\btheta}(\bx) &= \log \int\limits_{\bz^{p}} \int\limits_{\bz^{f}} p_{\btheta}(\bx \vert \bz^{p}, \bz^{f})~p(\bz^{p} \vert \bx)~p(\bz^{f} \vert \bx) \\ 
&= \log \int\limits_{\bz^{p}} \int\limits_{\bz^{f}} p_{\btheta}(\bx \vert \bz^{p}, \bz^{f})~p(\bz^{p} \vert \bx)~p(\bz^{f} \vert \bx)~\frac{q_{\bphi_{p}}(\bz^{p} \vert \bx)}{q_{\bphi_{p}}(\bz^{p} \vert \bx)}~\frac{q_{\bphi_{f}}(\bz^{f} \vert \bx)}{q_{\bphi_{f}}(\bz^{f} \vert \bx)} \nonumber \\
&= \log \mathbb{E}_{\substack{\bz^p \sim q_{\bphi_p} \\
                                   \bz^f \sim q_{\bphi_f}}} \frac{p_{\btheta}(\bx \vert \bz^{p}, \bz^{f})~p(\bz^{p} \vert \bx)~p(\bz^{f} \vert \bx) }{q_{\bphi_{p}}(\bz^{p} \vert \bx)~q_{\bphi_{f}}(\bz^{f} \vert \bx)} \nonumber \\
& \ge \mathbb{E}_{\substack{\bz^p \sim q_{\bphi_p} \\
                                   \bz^f \sim q_{\bphi_f}}} \log \frac{p_{\btheta}(\bx \vert \bz^{p}, \bz^{f})~p(\bz^{p} \vert \bx)~p(\bz^{f} \vert \bx) }{q_{\bphi_{p}}(\bz^{p} \vert \bx)~q_{\bphi_{f}}(\bz^{f} \vert \bx)} \nonumber \\
&= \mathbb{E}_{\substack{\bz^p \sim q_{\bphi_p} \\
                                   \bz^f \sim q_{\bphi_f}}} \log p_{\btheta}(\bx \vert \bz^{p}, \bz^{f}) - \mathbb{E}_{\bz^p \sim q_{\bphi_p}} \log \frac{q_{\bphi_{p}}(\bz^{p} \vert \bx)}{p(\bz^{p} \vert \bx)} - \mathbb{E}_{\bz^f \sim q_{\bphi_f}} \log \frac{q_{\bphi_{f}}(\bz^{f} \vert \bx)}{p(\bz^{f} \vert \bx)} \nonumber \nonumber \\
&= \mathbb{E}_{\substack{\bz^p \sim q_{\bphi_p} \\
                                  \bz^f \sim q_{\bphi_f}}} \log p_{\btheta}(\bx \vert \bz^{p}, \bz^{f}) - \KLDD{q_{\bphi_{p}}(\bz^{p} \vert \bx)}{p(\bz^{p} \vert \bx)} - \KLDD{q_{\bphi_{f}}(\bz^{f} \vert \bx)}{p(\bz^{f} \vert \bx)} \nonumber
\end{align}

We model the posterior distributions with two recurrent networks. The recurrent networks output two different posterior distributions, $q_{\bphi_p}(\bz^{p}_t \vert \bx_{1:t})$ and $q_{\bphi_f}(\bz^{f}_t \vert \bx_{1:t})$, at every time step. Due to the independence of the latent variables across time, $\bz^{p} = [\bz^{p}_1, \bz^{p}_2, \cdots, \bz^{p}_T]$ and $\bz^{f} = [\bz^{f}_1, \bz^{f}_2, \cdots, \bz^{f}_T]$, we can derive the estimation of posterior distributions across time steps as follows:
\begin{align}
    q_{\bphi_p}(\bz^p \vert \bx) &= \prod\limits_t q_{\bphi_p}(\bz^{p}_t \vert \bx_{1:t})  \nonumber \\
    q_{\bphi_f}(\bz^f \vert \bx) &= \prod\limits_t q_{\bphi_f}(\bz^{f}_t \vert \bx_{1:t})
\end{align}

Since the latent variables, $\bz^{p} = [\bz^{p}_1, \bz^{p}_2, \dots, \bz^{p}_T]$ and $\bz^{f} = [\bz^{f}_1, \bz^{f}_2, \dots, \bz^{f}_T]$, are independent across time and independent from each other, we can further decompose Kullback-Leibler terms in the evidence lower bound into individual time steps as in \eqnref{eq:slamp_baseline_kl_timesptes}.

At each time step, our model predicts $\bx_{t}$, conditioned on $\bx_{t-1}$, $\bz^{p}_{t}$, $\bz^{f}_{t}$. Since our model has recurrence connections, it considers not only $\bx_{t-1}$,  $\bz^{p}_{t}$ and $\bz^{f}_{t}$, but also $\bx_{1:t-2}$, $\bz^{p}_{1:t-1}$ and $\bz^{f}_{1:t-1}$. Therefore, we can further write our inference as:
\begin{align}
    \log p_{\btheta}(\bx \vert \bz^{p}, \bz^{f}) &= \log \prod\limits_t p_{\btheta}(\bx_t \vert \bx_{1:t-1}, \bz^{p}_{1:t}, \bz^{f}_{1:t})\\
    &= \sum\limits_t \log p_{\btheta}(\bx_t \vert \bx_{1:t-1}, \bz^{p}_{1:t}, \bz^{f}_{1:t}) \nonumber
\end{align}

Combining all of them leads to the following variational lower bound:
\begin{align}
\log p_{\btheta}(\bx) &\geq \mathcal{L}_{\btheta, \bphi_p, \bpsi_f}(\bx_{1:T}) \\
&=  \mathbb{E}_{\substack{\bz^p \sim q_{\bphi_p} \\
                                   \bz^f \sim q_{\bphi_f}}} \log p_{\btheta}(\bx \vert \bz^{p}, \bz^{f}) - \KLDD{q_{\bphi_p}(\bz^{p} \vert \bx)}{p_{\bpsi_p}(\bz^{p} \vert \bx)} - \KLDD{q_{\bphi_f}(\bz^{f} \vert \bx)}{p_{\bpsi_f}(\bz^{f} \vert \bx)} \nonumber \\
&=  \sum\limits_t \mathbb{E}_{\substack{\bz^p \sim q_{\bphi_p} \\
                                   \bz^f \sim q_{\bphi_f}}}   \log p_{\btheta}(\bx_t \vert \bx_{1:t-1}, \bz^{p}_{1:t}, \bz^{f}_{1:t}) \nonumber \\
& \hspace{15mm} -  \KLDD{q_{\bphi_p}(\bz^{p}_t \vert \bx_{1:t}}{p_{\bpsi_p}(\bz^{p}_t \vert \bx_{1:t-1})} \nonumber \\
& \hspace{15mm} - \KLDD{q_{\bphi_f}(\bz^{f}_t \vert \bx_{1:t}}{p_{\bpsi_f}(\bz^{f}_t \vert \bx_{1:t-1})} \nonumber
\end{align}
\section{Training Details}
\label{sec:training_details}
We provide training details including scheduled sampling (\secref{sec:sc_sampling}), architecture details (\secref{sec:arch_details}), and the hyper-parameters used in the optimization (\secref{sec:opt_params}).

\subsection{Scheduled Sampling}
\label{sec:sc_sampling}
Scheduled sampling proposed for sequence prediction \cite{Bengio2015NeurIPS} has been proven useful for several tasks where predictions need to be made based on the generated results from the previous time steps.
We also experiment with scheduled sampling as part of our training procedure.
Scheduled sampling prevents the model from conditioning on ground-truth perfect samples which are not available at test time. This is achieved by allowing the model to slowly encounter generated samples instead of ground-truth perfect samples. The ratio of ground-truth perfect samples over generated samples is decreased throughout the training.
Specifically, we apply inverse sigmoid decay. 
We report the scores with and without scheduled sampling for the proposed models, both SLAMP-Baseline and SLAMP, on all datasets. As can be seen from \tabref{tab:ablation}, scheduled sampling is not crucial but it improves the results on KTH, especially for SLAMP.

\begin{table}[h!]
    \caption{
        \label{tab:res-kth}
         \textbf{Ablation Study.} This table shows the quantitative results comparing SLAMP-Baseline and SLAMP with scheduled sampling (\textbf{+ SS}) and without during training.
         Following the previous work, we report the results as the mean and the $95\%$-confidence interval in terms of PSNR, SSIM, and LPIPS on all the datasets except LPIPS on MNIST.
    }
    \label{tab:ablation}
    \sisetup{detect-weight, table-align-uncertainty=true, mode=text}
    \renewrobustcmd{\bfseries}{\fontseries{b}\selectfont}
    \renewrobustcmd{\boldmath}{}
    \centering
    \vspace{0.1in}
    \begin{tabular}{clS[table-format=2.2(2)]S[table-format=1.4(2)]S[table-format=1.4(2)]}
        \toprule
        & Models & {PSNR} & {SSIM} & {LPIPS} \tabularnewline
        \midrule
        \multirow{4}{*}{\rotatebox[origin=c]{90}{\textsc{MNIST}}}
        & SLAMP-Baseline &  16.83 \pm 0.06 &  0.7537 \pm 0.0017 & {\textemdash} \tabularnewline
        & SLAMP-Baseline + SS & 16.32 \pm 0.06 & 0.7343 \pm 0.0016 & {\textemdash} \tabularnewline
        \cmidrule{2-5}
        & SLAMP  &  18.07 \pm 0.07 &  0.7736 \pm 0.0019 & {\textemdash} \tabularnewline
        & SLAMP + SS & 17.54 \pm 0.08 & 0.7567 \pm 0.0018 & {\textemdash} \tabularnewline
        \midrule \midrule
        \multirow{4}{*}{\rotatebox[origin=c]{90}{\textsc{KTH}}}
        & SLAMP-Baseline & 28.47 \pm 0.27 &  0.8527 \pm 0.0053 & 0.0896 \pm 0.0038 \tabularnewline
        & SLAMP-Baseline + SS & 29.20 \pm 0.28 &  0.8633 \pm 0.0048 & 0.0951 \pm 0.0036 \tabularnewline
        \cmidrule{2-5}
        & SLAMP  & 28.91 \pm 0.28 & 0.8604 \pm 0.0049 & 0.0860 \pm 0.0037 \tabularnewline
        & SLAMP + SS & 29.39 \pm 0.30 &  0.8646 \pm 0.0049 & 0.0795 \pm 0.0033 \tabularnewline
        \midrule \midrule
        \multirow{4}{*}{\rotatebox[origin=c]{90}{\textsc{BAIR}}}
        & SLAMP-Baseline & 19.60 \pm 0.26 &  0.8175 \pm 0.0083 & 0.0596 \pm 0.0031 \tabularnewline
        & SLAMP-Baseline + SS & 19.55 \pm 0.26 & 0.8171 \pm 0.0083 & 0.0634 \pm 0.0034 \tabularnewline
        \cmidrule{2-5}
        & SLAMP  & 19.67 \pm 0.26 &  0.8161 \pm 0.0086 & 0.0639 \pm 0.0037 \tabularnewline
        & SLAMP + SS & 19.75 \pm 0.26 &  0.8160 \pm 0.0084 & 0.0661 \pm 0.0035 \tabularnewline
        \bottomrule
    \end{tabular}
\end{table}

\subsection{Architecture Details}
\label{sec:arch_details}

\boldparagraph{Encoders and Decoders}
For all encoders and decoders, we use the same architectures as the previous work \cite{Denton2018ICML, Franceschi2020ICML}: a DCGAN \cite{Radford2016ICLR} generator and discriminator for MNIST, and a VGG16 architecture \cite{Simonyan2015ICLR} for KTH and BAIR datasets. In all datasets, we encode the image into an appearance feature vector of size $h_{appearance}=128$ and the two consecutive images into a motion feature vector of size $h_{motion}=128$. 
Compared to SVG, there are two more decoders for predicting flow and mask in our models. For flow decoder, we use the same decoder with two output channels representing motion in horizontal and vertical direction. See below for the details of the mask decoder.
For SLAMP, motion encoder takes concatenated frame pair as input and outputs a feature vector encoding motion from one frame to the next.

Similar to previous work \cite{Denton2018ICML, Franceschi2020ICML}, we  also use skip connections but with a minor modification. In the previous work, the skip connection from either the last conditioning frame or last generated frame is used. Instead, we take the running average of all the skip connections from seen or generated frames. For example, at time step 15, we use the average of previous 14 skip connections that are generated.

\boldparagraph{Mask Predictor}
For mask predictor, we use a 5-layer CNN with 2 Squeeze and Excitation layers (SE-Layer) \cite{hu2018CVPR} after each two convolutional layers. 
In the CNN, we use 64-channel filters at each layer and do not reduce the resolution by using $3\times3$ kernels with padding.
We simply concatenate the output of pixel decoder and warped prediction along their channel axis and feed it into mask predictor which outputs a one-channel image.
We apply sigmoid at the end to map the output to the range between 0 and 1, representing the weight to combine the appearance and the motion predictions.

\boldparagraph{LSTMs and Latent Variables}
For prior, posterior, and frame predictor LSTMs, we use the settings proposed in SVG \cite{Denton2018ICML}. All LSTMs have 256 neurons and all prior and posterior LSTMs have one layer whereas the frame predictor LSTMs have two layers.
For the size of the latent variables, we use $20$, $50$, $64$ for MNIST, KTH, and BAIR, respectively.

\subsection{Optimization Hyper-Parameters}
\label{sec:opt_params}
All the models are trained with Adam optimizer \cite{Kingma2015ICLR}, with decay rates $\beta_1=0.9$ and $\beta_2=0.999$. We train each model for 300 epochs where each epoch consists of 1000 updates, unless otherwise is specified. We take the model which performs the best in the validation set. We will share the trained models upon publication for replicating the results. Dataset-specific parameters for each dataset are as follows:

\boldparagraph{MNIST}
The batch size is chosen to be 32, learning rate is $3\times 10^{-4}$ and $\beta=1\times 10^{-4}$.
We continue training the models on MNIST with a lower learning rate, $1\times 10^{-5}$, a lower $\beta = 5\times 10^{-5}$, and a lower $\beta_1=0.7$.

\boldparagraph{KTH}
The batch size is chosen to be 20, learning rate is $1\times 10^{-4}$ and $\beta=1\times 10^{-6}$. We apply scheduled sampling with inverse sigmoid decay.

\boldparagraph{BAIR}
The batch size is chosen to be 20, learning rate is $1\times 10^{-4}$ and $\beta=1\times 10^{-4}$.

\boldparagraph{Training details for KITTI and Cityscapes}

We use $92\times310$ image resolution for KITTI and $128\times256$ for Cityscapes. We replaced LSTMs with ConvLSTMs and used $3\times10$ intermediate feature size for KITTI, $4\times8$ for Cityscapes. We used a shared encoder to downsample the image first and then, use two separate encoders for pixel and motion encoders to make model less powerful. We increased the number of layers in the shared encoder to downsample the higher resolution image, and preserve the VGG-basedd structure.

For SVG, we use the same settings as SLAMP. For SRVP, we use the same shared encoder and use a channel pooling at the end to make the convolutional feature vector compatible with the rest of the architecture.

We train all the models until the models see $2.4M$ video samples. We use the largest batch size that we could use and choose the learning rate $1\times10^{-4}$ for all the models.

\clearpage
\section{Detailed Quantitative Results}
\label{sec:quant_res}

In this section, we provide a detailed version of the quantitative results presented in the main paper in Figure 3 and 5. 

We compare the performance of SLAMP-Baseline and SLAMP to the previous work  in terms of PNSR, SSIM, and LPIPS averaged over all time steps on MNIST (\tabref{tab:res-mnist}), KTH (\tabref{tab:res-kth}), and BAIR (\tabref{tab:res-bair}) datasets. 
Confirming the results in the main paper, the proposed model SLAMP with motion history outperforms both the baseline model, SLAMP-Baseline, and the previous work \cite{Denton2018ICML, Babaeizadeh2018ICLR, Lee2018ARXIV} and performs comparably to the state of the art model SRVP \cite{Franceschi2020ICML}. See the main paper for a detailed analysis.

\begin{table}[h!]
    \caption{
        \label{tab:res-mnist}
        \textbf{Results on MNIST.} This table compares the results of SLAMP and SLAMP-Baseline to the previous work on MNIST dataset.
        Following the previous work, we report the results as the mean and the $95\%$-confidence interval in terms of PSNR and SSIM.
        Bold and underlined scores indicate the best and the second best performing method, respectively.
    }
    \sisetup{detect-weight, table-align-uncertainty=true, mode=text}
    \renewrobustcmd{\bfseries}{\fontseries{b}\selectfont}
    \renewrobustcmd{\boldmath}{}
    \centering
    \vspace{0.1in}
    \begin{tabular}{lcc}
        \toprule
        Models & {PSNR} & {SSIM} \tabularnewline
        \midrule
        SVG \cite{Denton2018ICML} & 14.50 $\pm$ 0.04 & 0.7090 $\pm$ 0.0015 \tabularnewline
        SRVP \cite{Franceschi2020ICML}& \underline{$16.93 \pm 0.07$} & \bfseries 0.7799 $\pm$ 0.0020 \tabularnewline
        SLAMP-Baseline & 16.83 $\pm$ 0.06 & 0.7537 $\pm$ 0.0018 \tabularnewline
        SLAMP & \bfseries 18.07 $\pm$ 0.08 &   \underline{$0.7736 \pm 0.0019$} \tabularnewline
        \bottomrule
    \end{tabular}
\end{table}

\begin{table}[h!]
    \caption{
        \label{tab:res-kth}
        \textbf{Results on KTH.} This table compares the results of SLAMP and SLAMP-Baseline to the previous work on KTH dataset.
        Following the previous work, we report the results as the mean and the $95\%$-confidence interval in terms of PSNR, SSIM, and LPIPS.
        Bold and underlined scores indicate the best and the second best performing method, respectively.
    }
    \sisetup{detect-weight, table-align-uncertainty=true, mode=text}
    \renewrobustcmd{\bfseries}{\fontseries{b}\selectfont}
    \renewrobustcmd{\boldmath}{}
    \centering
    \vspace{0.1in}
    \begin{tabular}{lccc}
        \toprule
        Models & {PSNR} & {SSIM} & {LPIPS} \tabularnewline
        \midrule
        SV2P \cite{Babaeizadeh2018ICLR} & 28.19 $\pm$ 0.31 & 0.8141 $\pm$ 0.0050 & 0.2049 $\pm$ 0.0053 \tabularnewline
        SAVP \cite{Lee2018ARXIV} & 26.51 $\pm$ 0.29 & 0.7564 $\pm$ 0.0062 & 0.1120 $\pm$ 0.0039 \tabularnewline
        SVG \cite{Denton2018ICML} & 28.06 $\pm$ 0.29 & 0.8438 $\pm$ 0.0054 & 0.0923 $\pm$ 0.0038 \tabularnewline
        SRVP \cite{Franceschi2020ICML} & \bfseries 29.69 $\pm$ 0.32 & \bfseries 0.8697 $\pm$ 0.0046 & \bfseries 0.0736 $\pm$ 0.0029 \tabularnewline
        SLAMP-Baseline &  29.20 $\pm$ 0.28 & 0.8633 $\pm$ 0.0048 & 0.0951 $\pm$ 0.0036 \tabularnewline
        SLAMP & \underline{$29.39 \pm 0.30$} & \underline{$0.8646 \pm 0.0050$} & \underline{$0.0795 \pm 0.0034$} \tabularnewline
        \bottomrule
    \end{tabular}
\end{table}

\begin{table}[h!]
    \caption{
        \label{tab:res-bair}
        \textbf{Results on BAIR.} This table compares the results of SLAMP and SLAMP-Baseline to the previous work on BAIR dataset.
        Following the previous work, we report the results as the mean and the $95\%$-confidence interval in terms of PSNR, SSIM, and LPIPS.
        Bold and underlined scores indicate the best and the second best performing method, respectively.
    }
    \sisetup{detect-weight, table-align-uncertainty=true, mode=text}
    \renewrobustcmd{\bfseries}{\fontseries{b}\selectfont}
    \renewrobustcmd{\boldmath}{}
    \centering
    \vspace{0.1in}
    \begin{tabular}{lccc}
        \toprule
        Models & {PSNR} & {SSIM} & {LPIPS} \tabularnewline
        \midrule
        SV2P \cite{Babaeizadeh2018ICLR}& \bfseries 20.39 $\pm$ 0.27 &  0.8169 $\pm$ 0.0086 & 0.0912 $\pm$ 0.0053 \tabularnewline
        SAVP \cite{Lee2018ARXIV} & 18.44 $\pm$ 0.25 & 0.7887 $\pm$ 0.0092 & 0.0634 $\pm$ 0.0026 \tabularnewline
        SVG \cite{Denton2018ICML}& 18.95 $\pm$ 0.26 & 0.8058 $\pm$ 0.0088 & 0.0609 $\pm$ 0.0034 \tabularnewline
        SRVP \cite{Franceschi2020ICML} & 19.59 $\pm$ 0.27 & \bfseries 0.8196 $\pm$ 0.0084 & \bfseries 0.0574 $\pm$ 0.0032 \tabularnewline
        SLAMP-Baseline & 19.60 $\pm$ 0.26 &  \underline{$0.8175 \pm 0.0084$} &  \underline{$0.0596 \pm 0.0032$} \tabularnewline
        SLAMP & \underline{$19.67 \pm 0.26$} &  0.8161 $\pm$ 0.0086 & 0.0639 $\pm$ 0.0037 \tabularnewline
        \bottomrule
    \end{tabular}
\end{table}

In addition, we provide detailed results corresponding to the components of our model. We evaluate the result of the static head, the dynamic head, and simply their average without the learned mask and compare them to our full model with the learned mask in \tabref{tab:mask_ablation}. Our full model, SLAMP, performs the best in all datasets according to all three evaluation metrics by combining the two predictions according to the mask prediction.

\begin{table}[t]
    \caption{
         \textbf{Mask Ablation Study.} This table shows the quantitative results comparing the components of SLAMP model. \textbf{Static} refers to the evaluation of the prediction of the static head directly, \textbf{Dynamic} refers to the evaluation of the prediction of the dynamic head which uses optical flow to predict the next frame, \textbf{Average} refers to average of static and dynamic heads without using the mask prediction. The last row for each dataset show the results of our model which uses the mask prediction to fuse the static and dynamic predictions.
         Following the previous work, we report the results as the mean and the $95\%$-confidence interval in terms of PSNR, SSIM, and LPIPS on all the datasets except LPIPS on MNIST.
    }
    \label{tab:mask_ablation}
    \sisetup{detect-weight, table-align-uncertainty=true, mode=text}
    \renewrobustcmd{\bfseries}{\fontseries{b}\selectfont}
    \renewrobustcmd{\boldmath}{}
    \centering
    \vspace{0.1in}
    \begin{tabular}{clccc}
        \toprule
        & Models & {PSNR} & {SSIM} & {LPIPS} \tabularnewline
        \midrule
        \multirow{4}{*}{\rotatebox[origin=c]{90}{\textsc{MNIST}}}
        & SLAMP - Static &  15.03 $\pm$ 0.04 &  0.7273 $\pm$ 0.0014 & {\textemdash} \tabularnewline
        & SLAMP - Dynamic & 17.64 $\pm$ 0.09 & 0.7639 $\pm$ 0.0019 & {\textemdash} \tabularnewline
        & SLAMP - Average &  15.96 $\pm$ 0.05 &  0.7377 $\pm$ 0.0017 & {\textemdash} \tabularnewline
        & SLAMP & \bfseries 18.07 $\pm$ 0.07 & \bfseries  0.7736 $\pm$ 0.0019 & {\textemdash} \tabularnewline
        \midrule \midrule
        \multirow{4}{*}{\rotatebox[origin=c]{90}{\textsc{KTH}}}
        & SLAMP - Static & 28.12 $\pm$ 0.28 &  0.8410 $\pm$ 0.0056 & 0.0844 $\pm$ 0.0037 \tabularnewline
        & SLAMP - Dynamic & 16.14 $\pm$ 0.14 &  0.7614 $\pm$ 0.0064 & 0.3689 $\pm$ 0.0080 \tabularnewline
        & SLAMP - Average  & 21.61 $\pm$ 0.11 & 0.8359 $\pm$ 0.0048 & 0.2039 $\pm$ 0.0054 \tabularnewline
        & SLAMP & \bfseries 29.39 $\pm$ 0.30 &  \bfseries 0.8646 $\pm$ 0.0049 & \bfseries 0.0795 $\pm$ 0.0033 \tabularnewline
        \midrule \midrule
        \multirow{4}{*}{\rotatebox[origin=c]{90}{\textsc{BAIR}}}
        & SLAMP - Static & 19.38 $\pm$ 0.26 &  0.8119 $\pm$ 0.00837 & 0.0606 $\pm$ 0.0033 \tabularnewline
        & SLAMP - Dynamic & 16.65 $\pm$ 0.20 & 0.7643 $\pm$ 0.0010 & 0.1176 $\pm$ 0.0048 \tabularnewline
        & SLAMP - Average  & 18.51 $\pm$ 0.20 &  0.8073 $\pm$ 0.0088 & 0.0897 $\pm$ 0.0044 \tabularnewline
        & SLAMP & \bfseries 19.75 $\pm$ 0.26 &  \bfseries 0.8160 $\pm$ 0.0084 & \bfseries 0.0661 $\pm$ 0.0035 \tabularnewline
        \bottomrule
    \end{tabular}
    \vspace*{5in}
\end{table}

\clearpage
\section{Additional Visualizations and Qualitative Results}
\label{sec:add_qual}

\subsection{Optical Flow Visualization}
\begin{wrapfigure}{R}{0.25\textwidth}
\vspace{-0.5cm}
\centering
\includegraphics[width=0.15\textwidth]{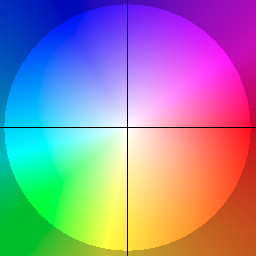}
\caption{\textbf{Optical Flow False Coloring}. Colors on the wheel indicate the direction of motion in 2D.}
\label{fig:color_wheel}
\end{wrapfigure}
\figref{fig:color_wheel} shows the color wheel used to visualize the optical flow with false coloring. Colors show the direction of motion and the intensity of color in the visualizations show the magnitude of motion, \ie intense colors for large motions. By following the usual practice in optical flow, we predict flow from the target frame to the current frame and apply inverse warping to obtain the target frame. Therefore, the direction of motion is also inverse, \ie the opposite direction on the wheel shows the motion from the current frame to the target frame.

\subsection{Comparison of Static and Dynamic Latent Variables}
\begin{figure}[H]
\centering
\includegraphics[width=0.8\textwidth]{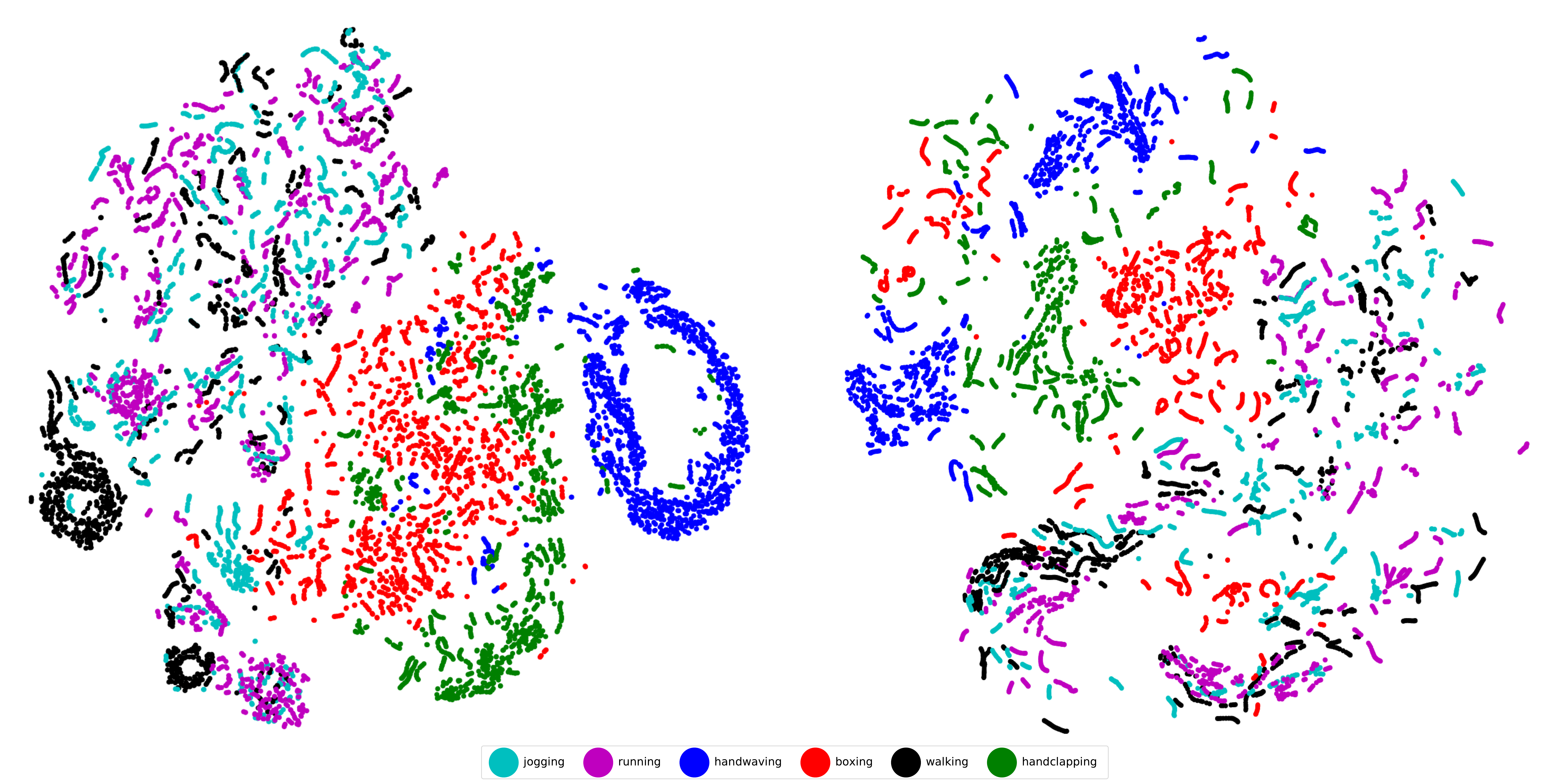}
\caption{\textbf{Dynamic (left) vs. Static (right) Latent Variables}. This figure shows the T-SNE visualization of dynamic and static latent variables on 300 test videos from KTH dataset. In dynamic latent variables, different classes with similar repetitive movements such as walking, running, and jogging are clustered together. However, in static latent variables, points are more scattered and do not form clusters according to semantic actions.}
\label{fig:tsne-both}
\end{figure}

In Fig. 8 of the main paper, we provide a visualization of stochastic latent variables of the dynamic component on KTH using t-SNE. Here, we provide both the static and the dynamic components for a comparison. The same colors from the main paper show the semantic classes of video frames plotted. As can be seen from \figref{fig:tsne-both}, static variables on the right are more scattered and do not from clusters according to semantic classes as in the dynamic variables on the left (and in the main paper). This shows that our model can learn video dynamics according to semantic classes with separate modelling of the dynamic component.

\begin{figure}[H]
\centering
\includegraphics[width=0.5\textwidth]{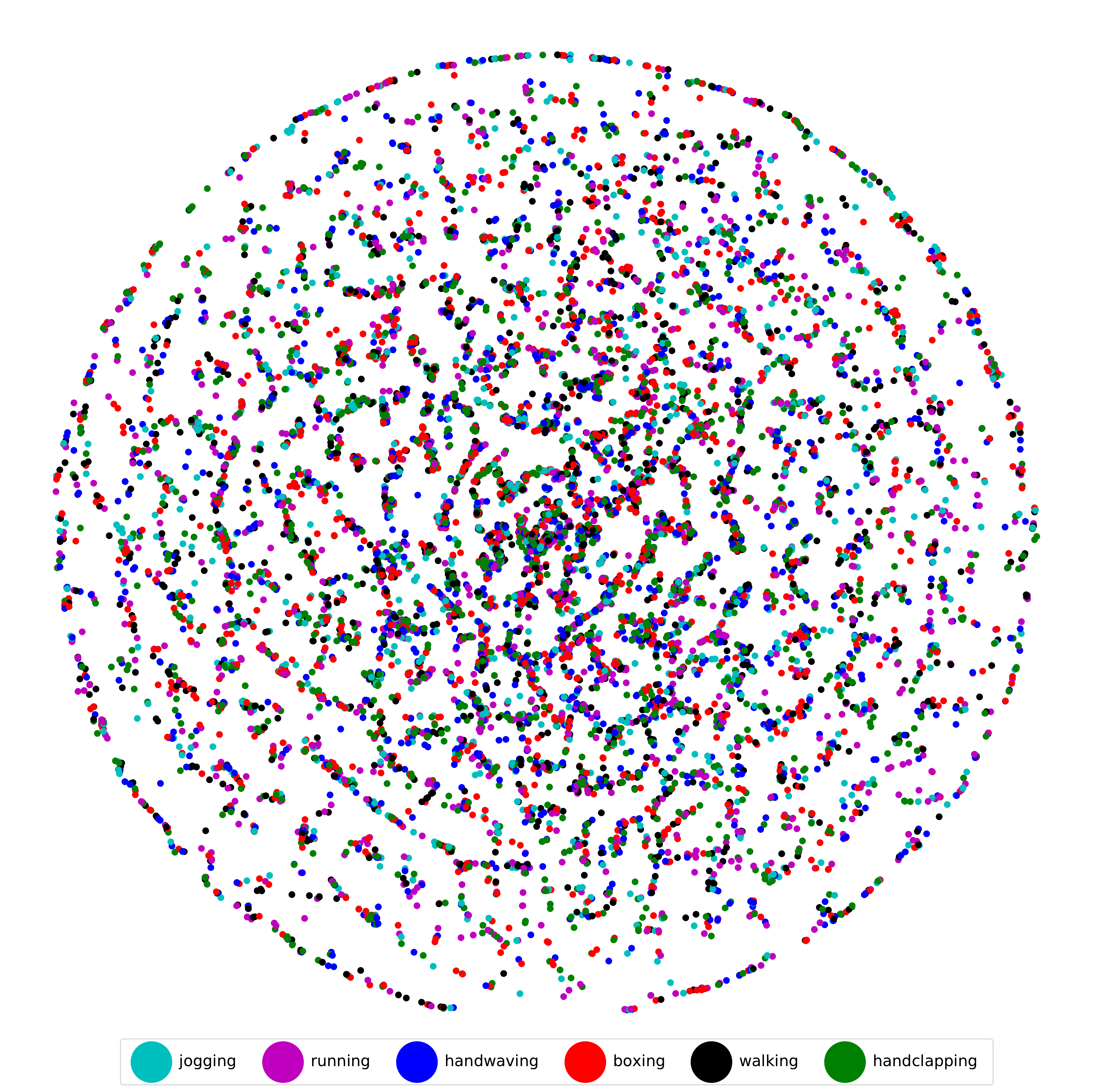}
\caption{\textbf{Latent Variables of SVG}. This figure shows the T-SNE visualization of latent variables of SVG \cite{Denton2018ICML} on 300 test videos from KTH dataset. The latent variables form a normal distribution when visualized with T-SNE because SVG method uses a standard normal distribution as a fixed prior in KTH dataset. The model learns the general image information instead of motions groupings.}
\label{fig:tsne-svg}
\end{figure}

In \figref{fig:tsne-svg}, we visualize the latent variables of SVG \cite{Denton2018ICML} to show the difference between our architecture's latent variables and SVG's latent variables. In our architecture, dynamic branch learns the similar repetitive movements whereas static branch learns the general image information. Therefore, dynamic latent variables form cluster around similar movements. However, in SVG, there is only one branch to predict the future frames, which only encodes the general image information rather than motion cues. Therefore, samples do not form clusters according to semantic classes as in our case for dynamic latent variables shown on the left in \figref{fig:tsne-both}.

\subsection{Diversity of Generated Samples}
\label{sec:diversity}
\begin{figure}[H]
\centering
\includegraphics[width=\textwidth]{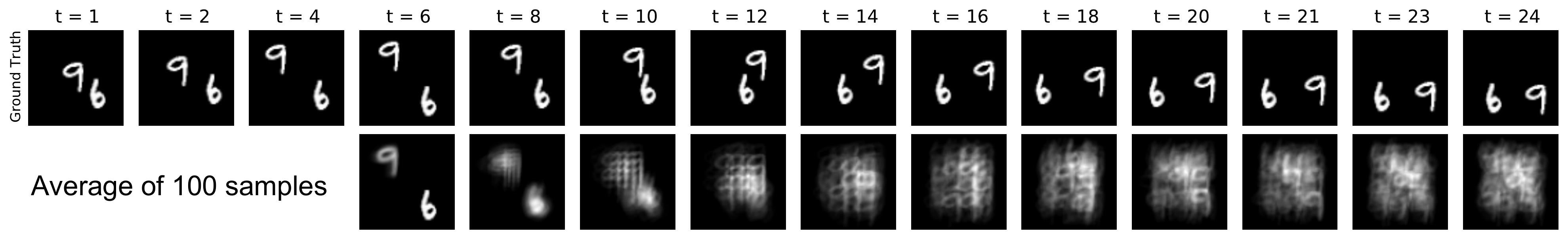}
\caption{\textbf{Diversity on MNIST.} After a digit hits the wall, it can move in any direction. Our model successfully models the stochasticity of this case and generates diverse results, resulting in blurry average images.}
\label{fig:diversity_mnist}
\end{figure}

\begin{figure}[H]
\centering
\includegraphics[width=\textwidth]{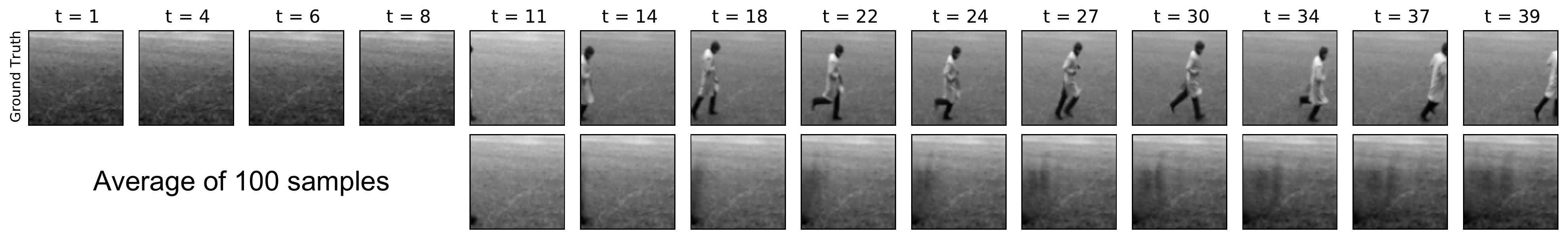}
\caption{\textbf{Diversity on KTH.} Since the running person appears after the conditioning frames, the model should generate different results for each sample. The average of the generated samples does not contain any human because our model can generate diverse results, \eg person in various poses appearing at different time steps with different speed.}
\label{fig:diversity_kth}
\end{figure}

\begin{figure}[H]
\centering
\includegraphics[width=\textwidth]{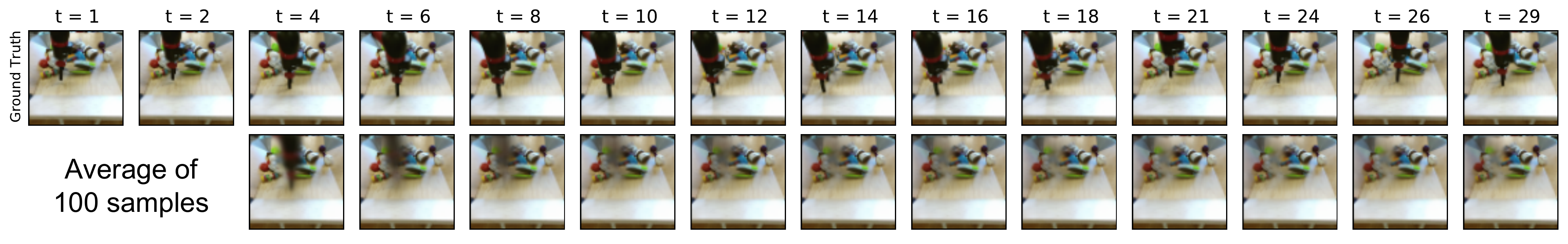}
\caption{\textbf{Diversity on BAIR.} The robot hand can move in any direction at each time step, therefore the generated samples should differ from each other in terms of the position of the robot hand. The moving robot hand becomes invisible in the average images after the first few frames, which is an indication of the diversity of the generated samples.}
\label{fig:diversity_bair}
\end{figure}

As proposed in SAVP \cite{Lee2018ARXIV}, as a measure of diversity, we visualize the average over 100 generated samples. According to this measure, if a model is able to generate diverse results, generated samples should differ where there is motion, \eg a moving object appearing in different positions and moving in different directions, leading to blurring out of the moving object. Therefore, we expect to see the background without moving objects in the average of the generated samples. The average samples confirm this for our model as shown for MNIST \figref{fig:diversity_mnist}, KTH \figref{fig:diversity_kth}, and BAIR \figref{fig:diversity_bair}.

\begin{figure}[H]
\centering
\includegraphics[width=\textwidth]{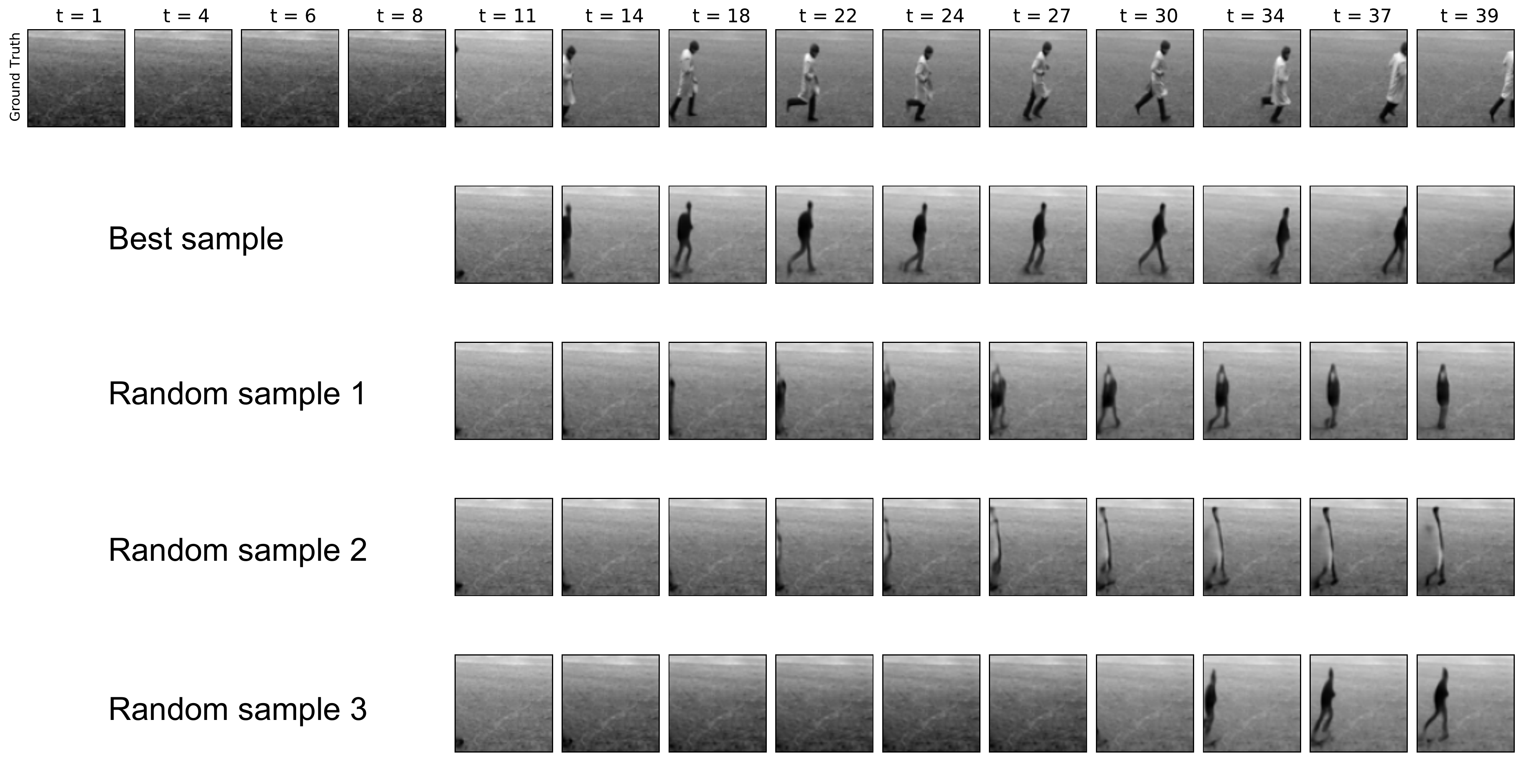}
\caption{\textbf{Subject Appearing after Conditioning Frames on KTH.} We show the best sample and three random samples for the case where the subject appears after conditioning frames. Our model generates different results at each random sample by learning dataset dynamics.}
\label{fig:diversity_kth_no_person}
\end{figure}

There is a special case on KTH which further supports our diversity claim as shown in Fig. 7 of the main paper. When subject appears after conditioning frames, our model can handle stochasticity of this challenging case and can generate \textbf{diverse} sequences. We show the best prediction and three random predictions in \figref{fig:diversity_kth_no_person}. Generated samples differ in terms of pose and speed of the subject as well as the time step that the subject appears.

\subsection{Additional Qualitative Results}
\label{sec:add_qualitative}
For each dataset, we show random examples with details of the best sample as well as three random samples generated. The detailed visualizations show appearance and motion prediction separately as well as the mask prediction and optical flow with false coloring. Random sample visualizations show the best sample and three random samples. We also provide full sequences of the samples in \figref{fig:teaser} in \figref{fig:full_seq_teaser_city} and \ref{fig:full_seq_teaser_kitti}.

%
\begin{figure}[H]
\centering
\includegraphics[width=\textwidth]{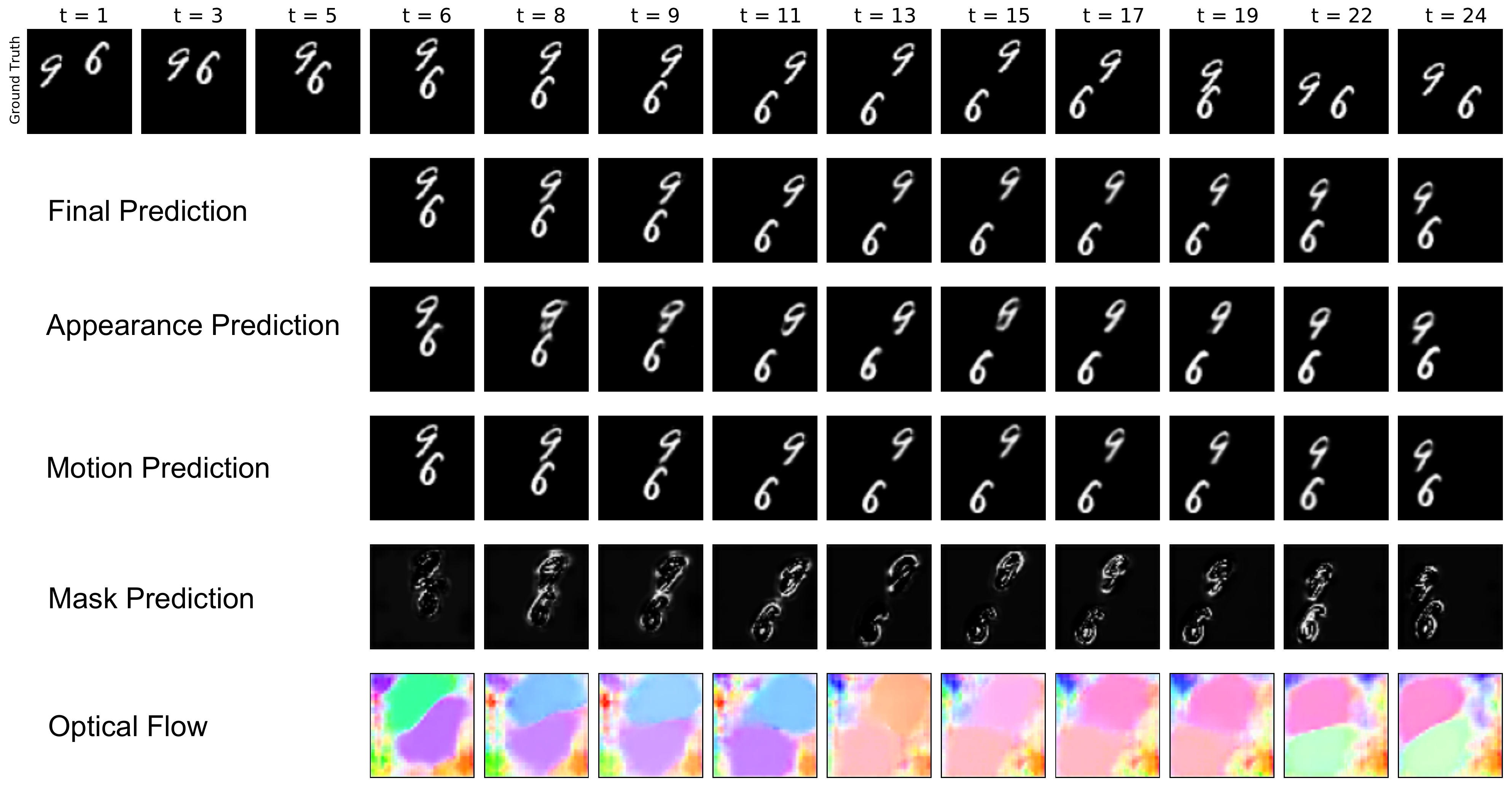}
\caption{\textbf{No Overlapping Digits.} This figure show a regular case with two non-overlapping digits. Note that predicted flow is from the target frame to the current frame since we apply inverse warping. The correctness of the optical flow estimation can be verified by inspecting \figref{fig:color_wheel}.}
\label{fig:mnist_no_overlap}
\end{figure}
\begin{figure}[H]
\centering
\includegraphics[width=\textwidth]{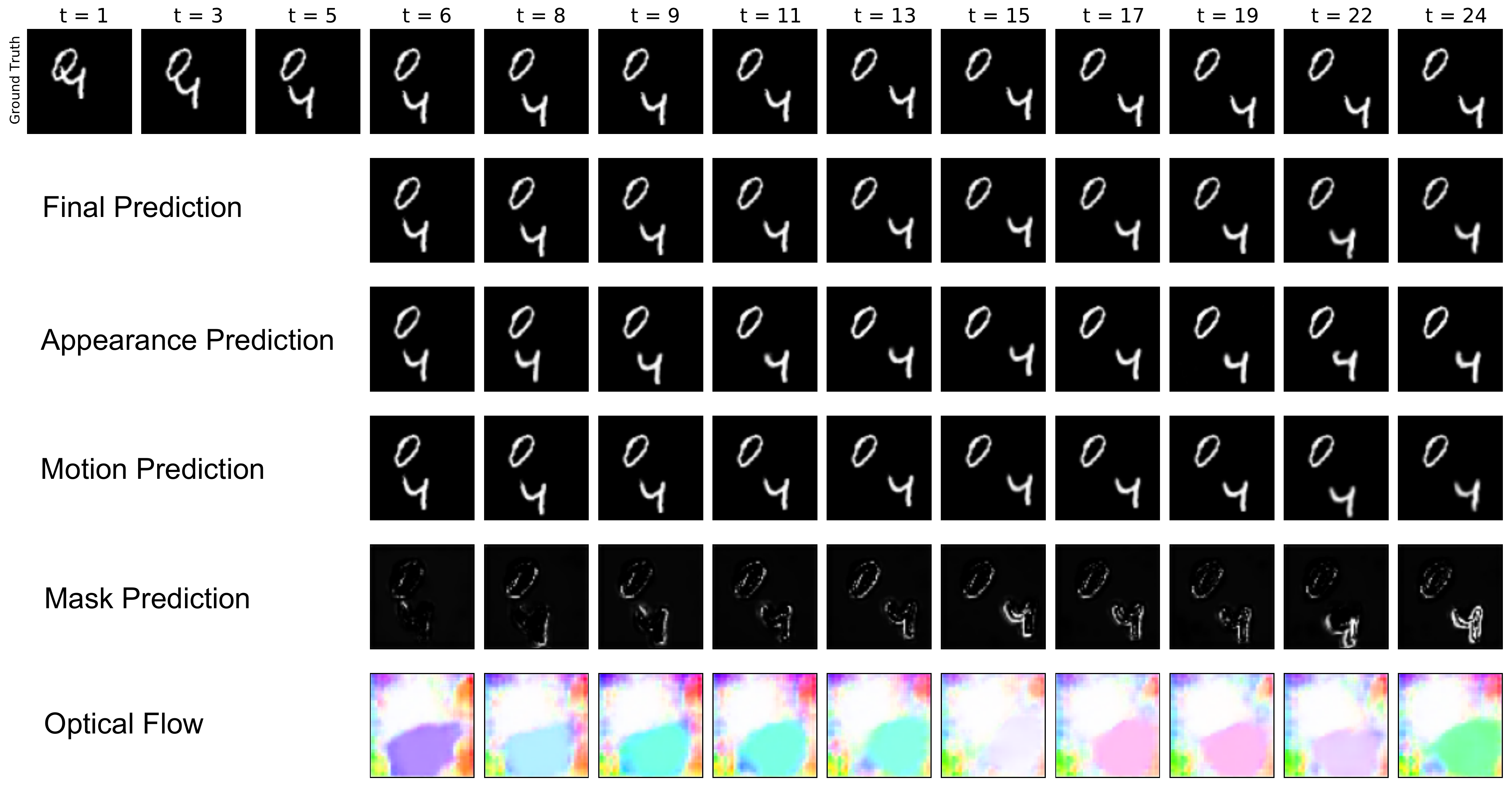}
\caption{\textbf{Stationary Digit.} This figure shows a case where a digit, \ie 0, is not moving. As can be seen from the last row, optical flow is correctly estimated as zero for that digit.}
\label{fig:mnist_no_motion}
\end{figure}
\begin{figure}[H]
\centering
\includegraphics[width=\textwidth]{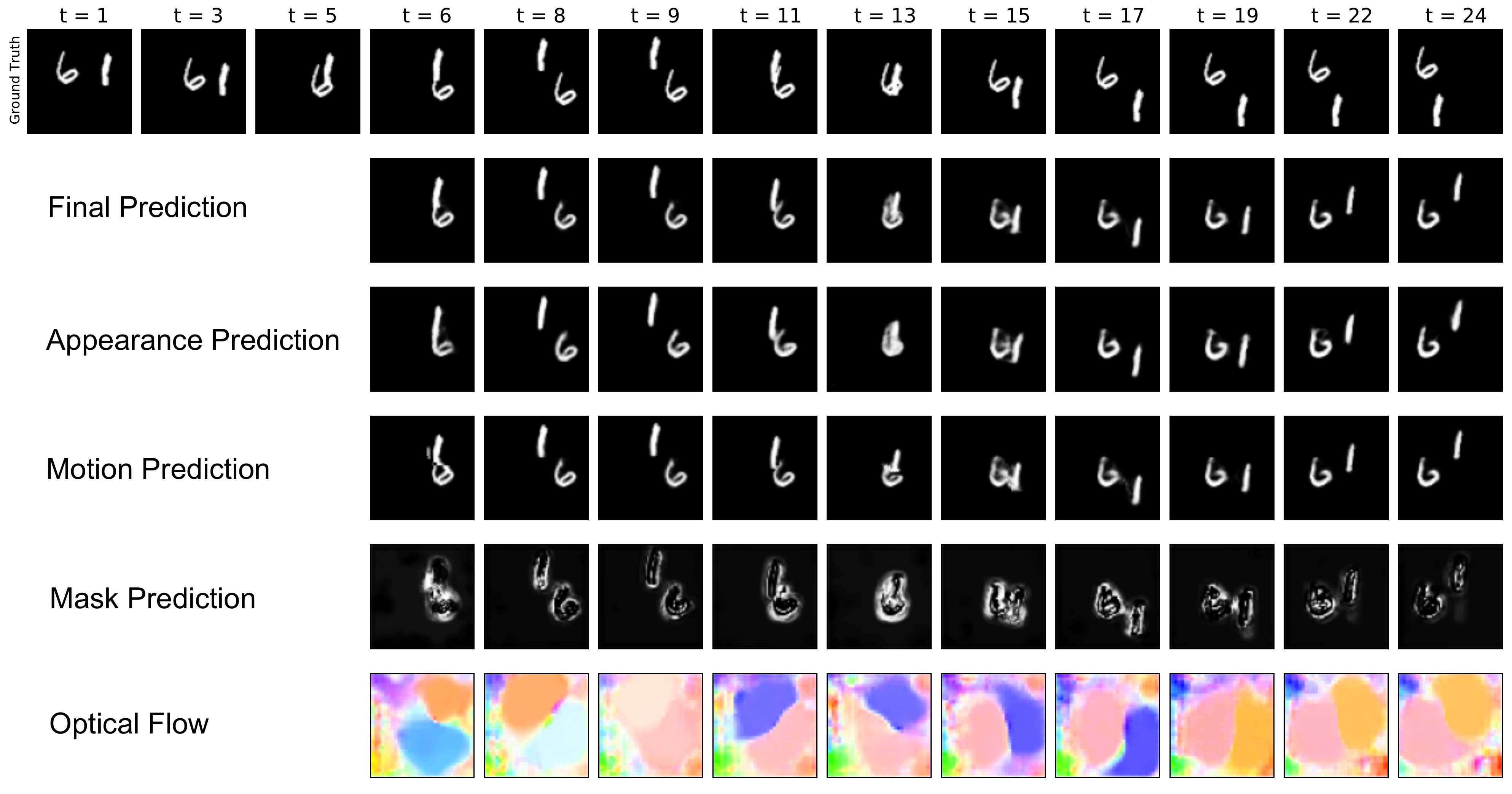}
\caption{\textbf{Overlapping Digits.} This figure shows a challenging case where two digits cross each other and continue moving. The digits start overlapping at around $t=13$. Our model can successfully handle this challenging case by preserving the appearance of digits.}
\label{fig:mnist_overlap1}
\end{figure}
\begin{figure}[H]
\centering
\includegraphics[width=\textwidth]{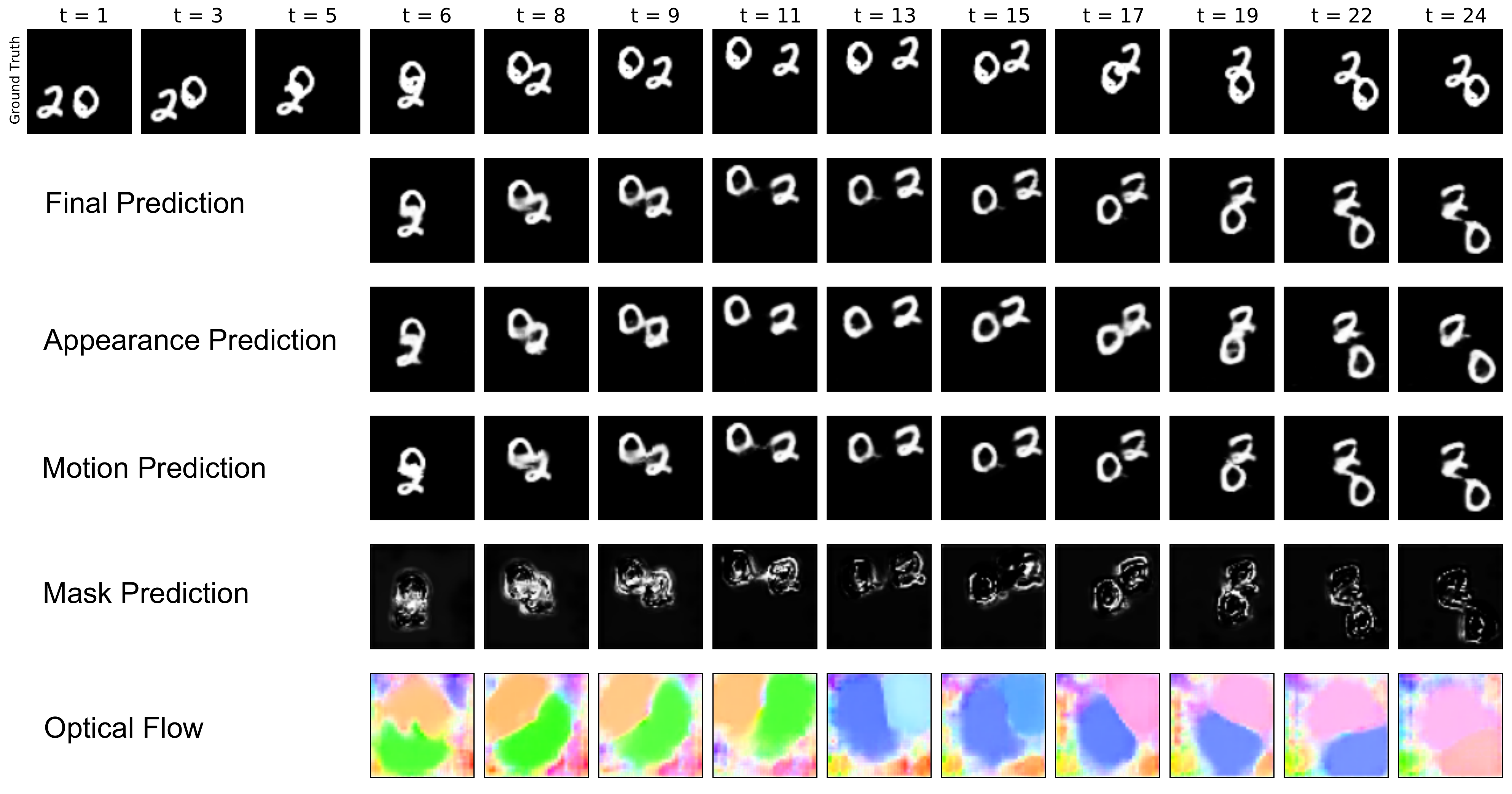}
\caption{\textbf{Overlapping Digits.} This figure shows a challenging case where two digits cross each other and continue moving. The digits start overlapping at around $t=6$. Our model can successfully handle this challenging case by preserving the appearance of digits.}
\label{fig:mnist_overlap2}
\end{figure}
\begin{figure}[H]
\centering
\includegraphics[width=\textwidth]{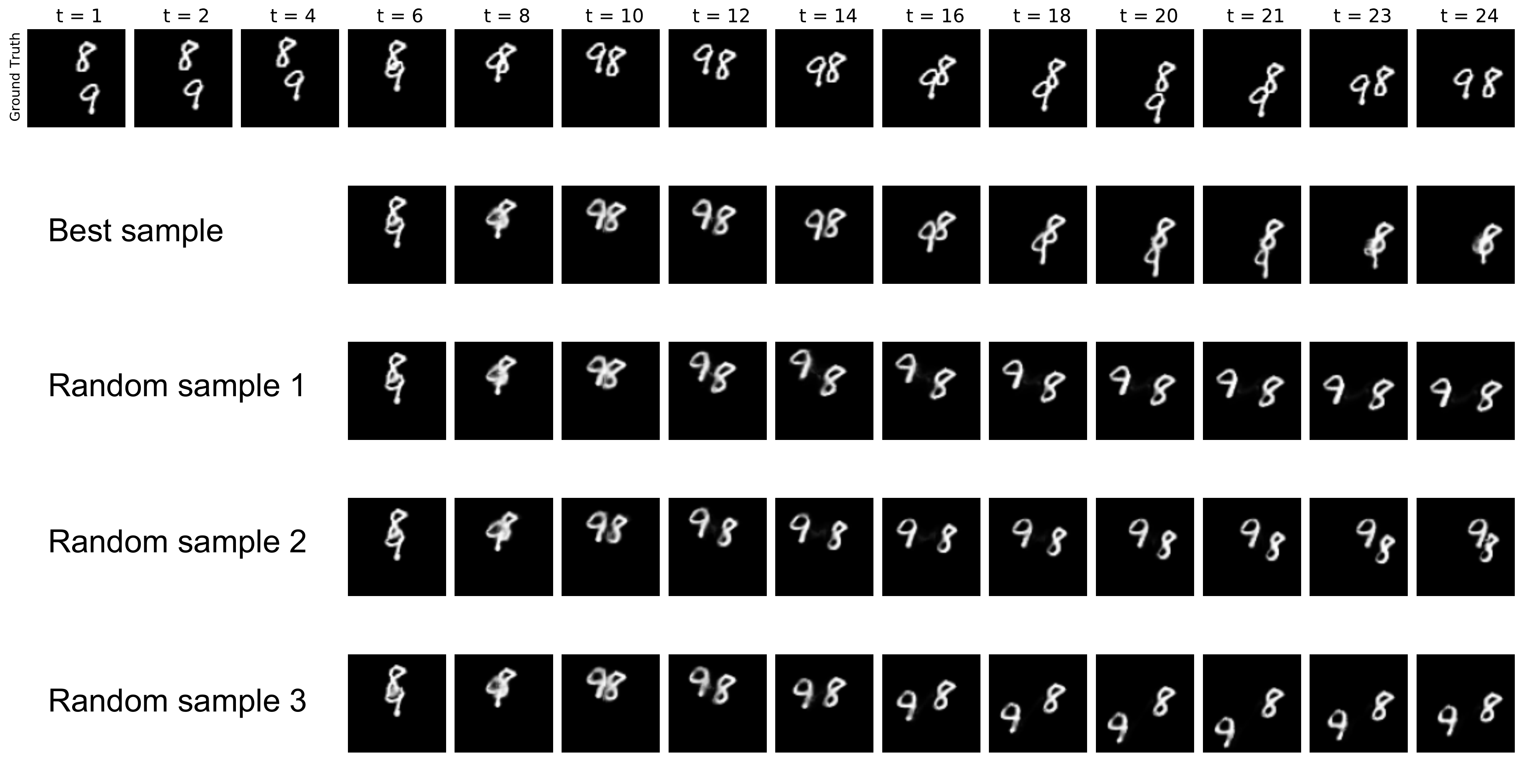}
\caption{\textbf{Random Samples.} We show the best sample and three random samples generated. All of the predictions are sharp-looking and different than each other, which proves that our model can generate diverse results.}
\label{fig:mnist_random1}
\end{figure}
\begin{figure}[H]
\centering
\includegraphics[width=\textwidth]{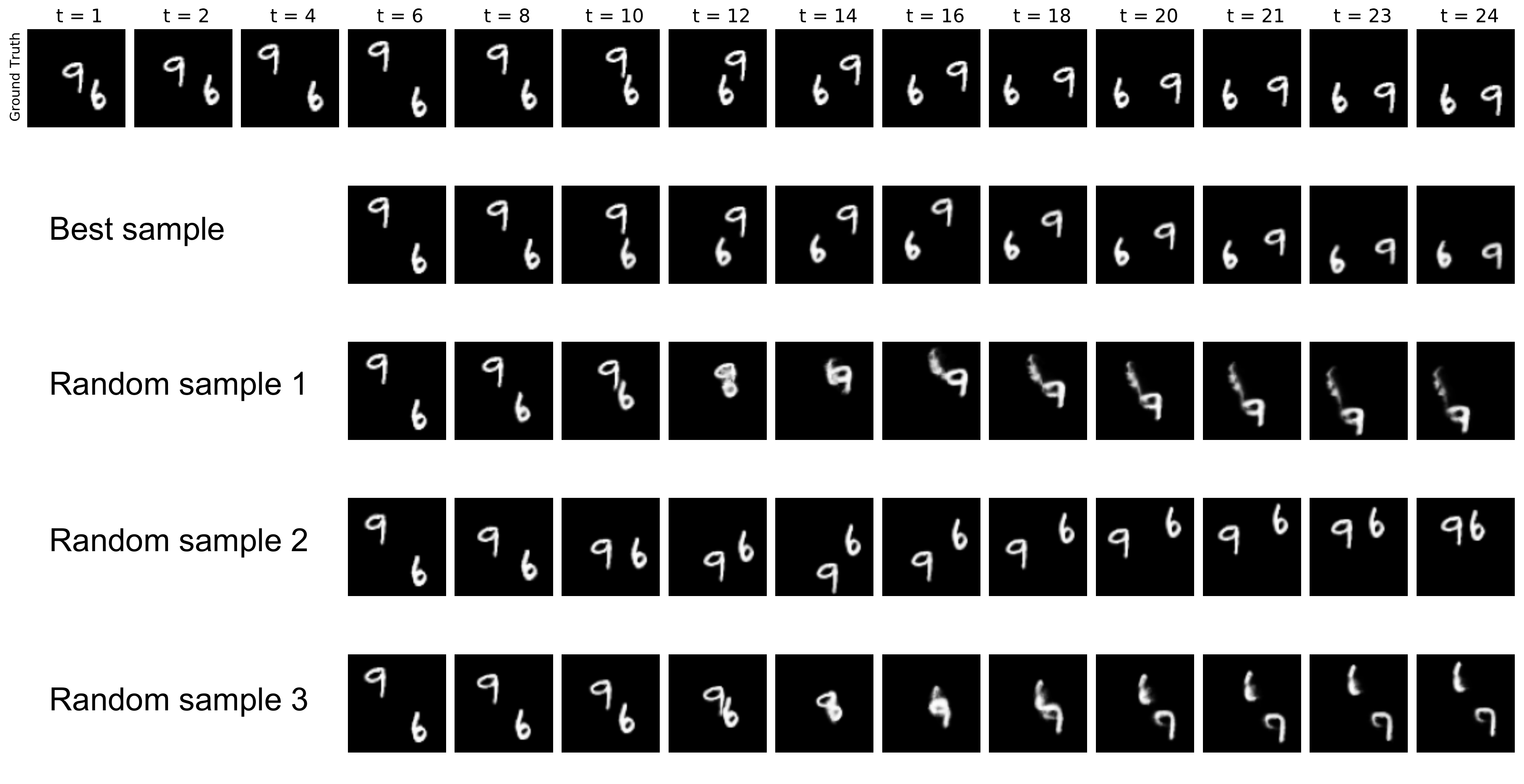}
\caption{\textbf{Random Samples.} We show the best sample and three random samples generated. The first and the third samples cannot preserve the shape of the digits, however, the best sample and the second sample are still sharp-looking.}
\label{fig:mnist_random2}
\end{figure}

\begin{figure}[H]
\centering
\includegraphics[width=\textwidth]{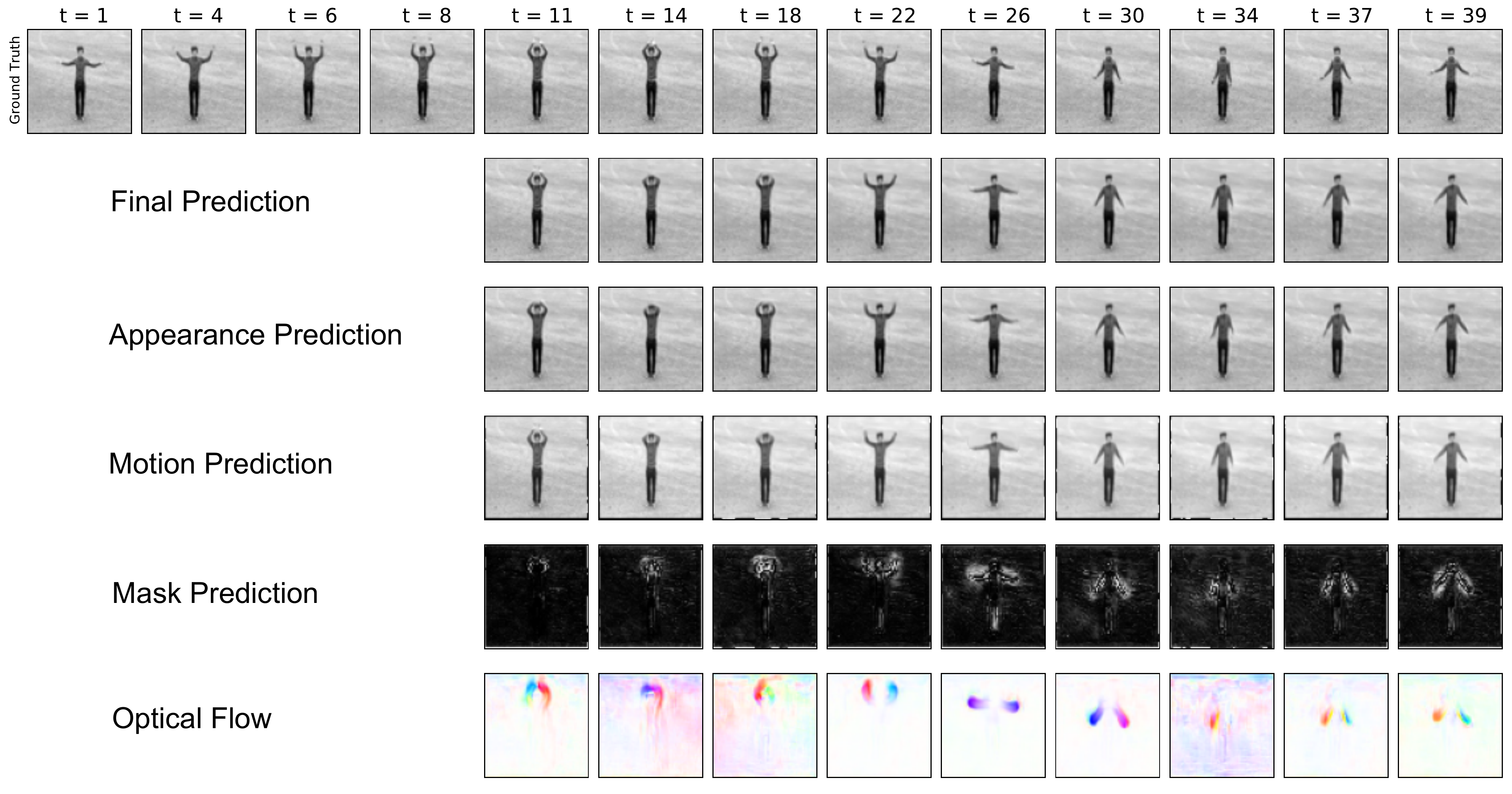}
\caption{\textbf{Person Waving Hands.} This figure shows the challenging case of hand waving with fast motion in a small region. Note that predicted flow is from the target frame to the current frame since we apply inverse warping. The correctness of the optical flow estimation can be verified by checking \figref{fig:color_wheel}. Our model focuses on the motion prediction for moving hands but it recovers the occluded motion boundaries from the appearance prediction.}
\label{fig:kth_waving}
\end{figure}
\begin{figure}[H]
\centering
\includegraphics[width=\textwidth]{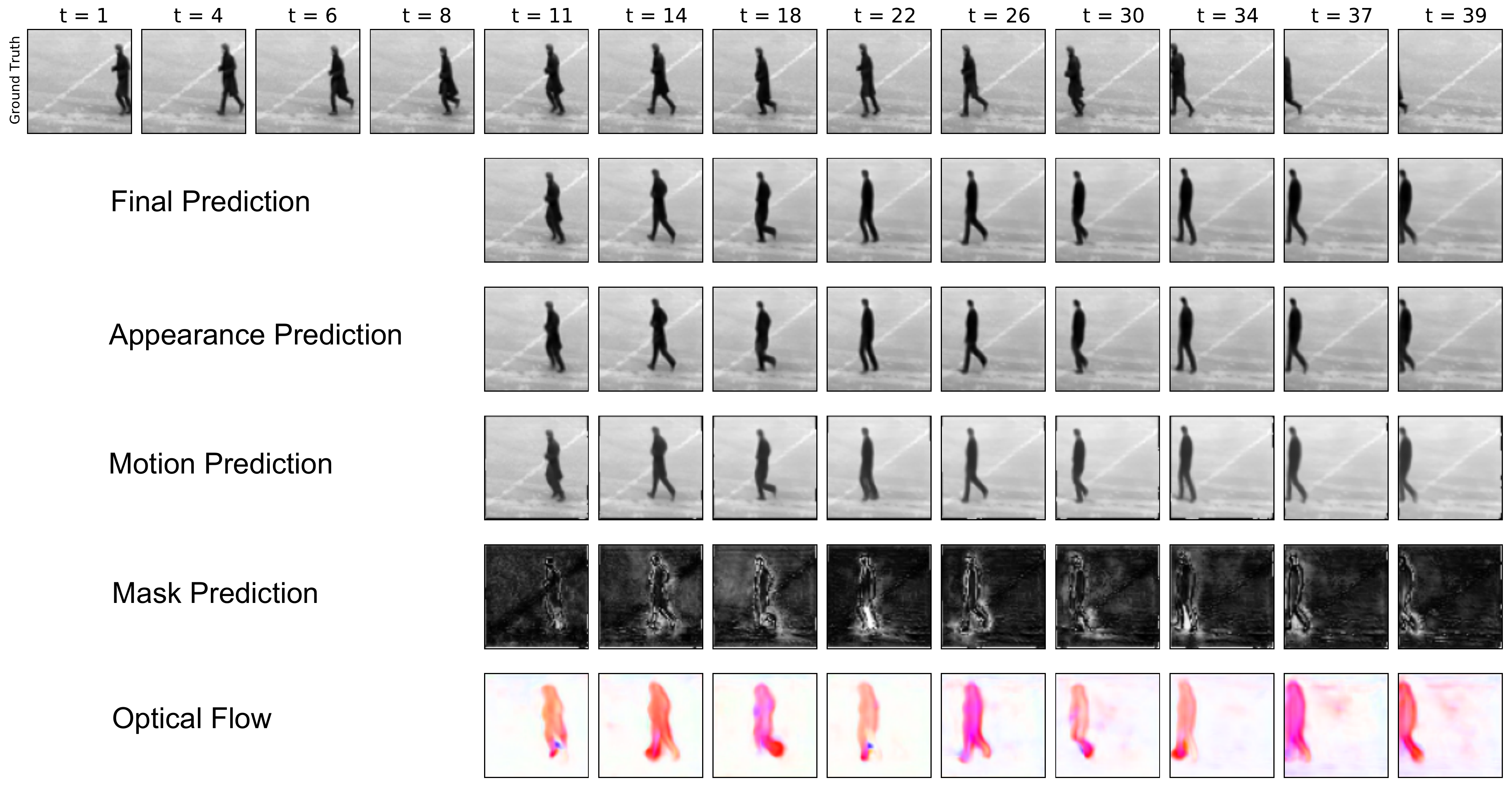}
\caption{\textbf{Person Running.} Our model correctly estimates the optical flow showing the motion of a person running and it can recover the occluded pixels around the legs from the appearance prediction.}
\label{fig:kth_running1}
\end{figure}
\begin{figure}[H]
\centering
\includegraphics[width=\textwidth]{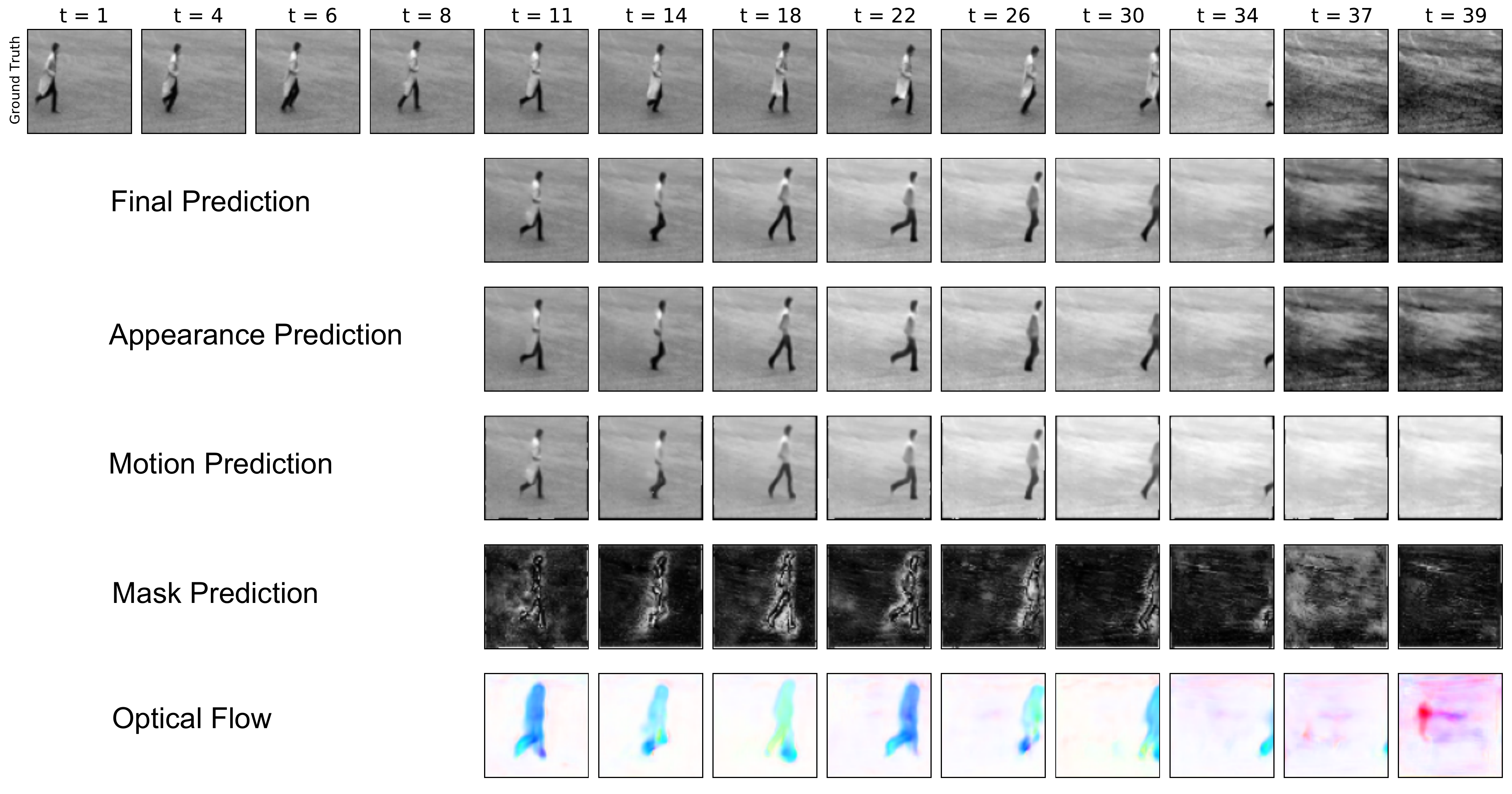}
\caption{\textbf{Person Leaving the Frame.} Our model correctly estimates optical flow when the person leaves the scene in the middle of the sequence by predicting nearly zero flow towards the end.}
\label{fig:kth_running2}
\end{figure}
\begin{figure}[H]
\centering
\includegraphics[width=\textwidth]{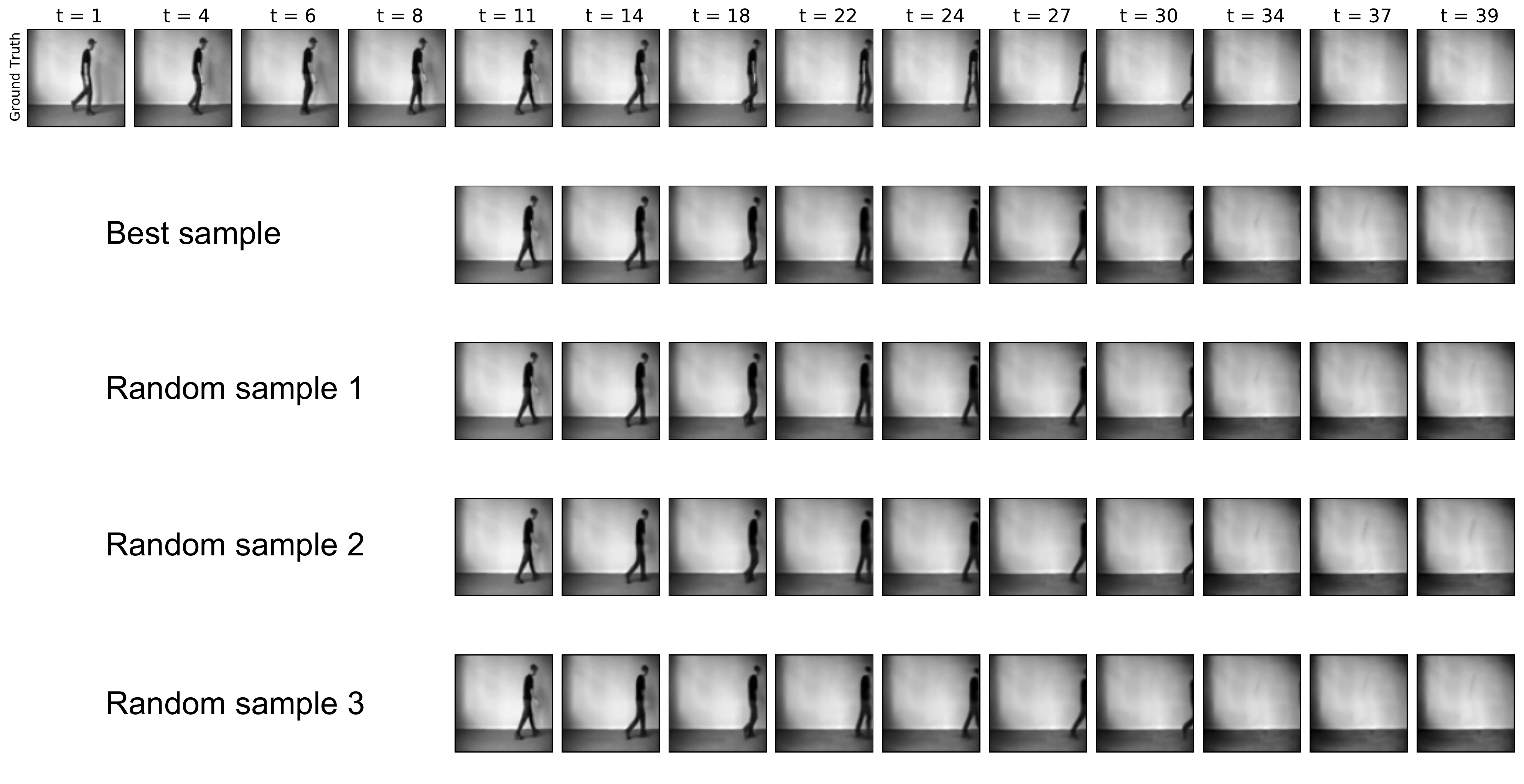}
\caption{\textbf{Random Samples.} We show the best sample and three random samples generated. Random samples look very similar due to the regular motion in the conditioning frames. Our model can capture the motion from the conditioning frames by generating consistent samples, only with minor differences in speed.}
\label{fig:kth_random}
\end{figure}


\begin{figure}[H]
\centering
\includegraphics[width=\textwidth]{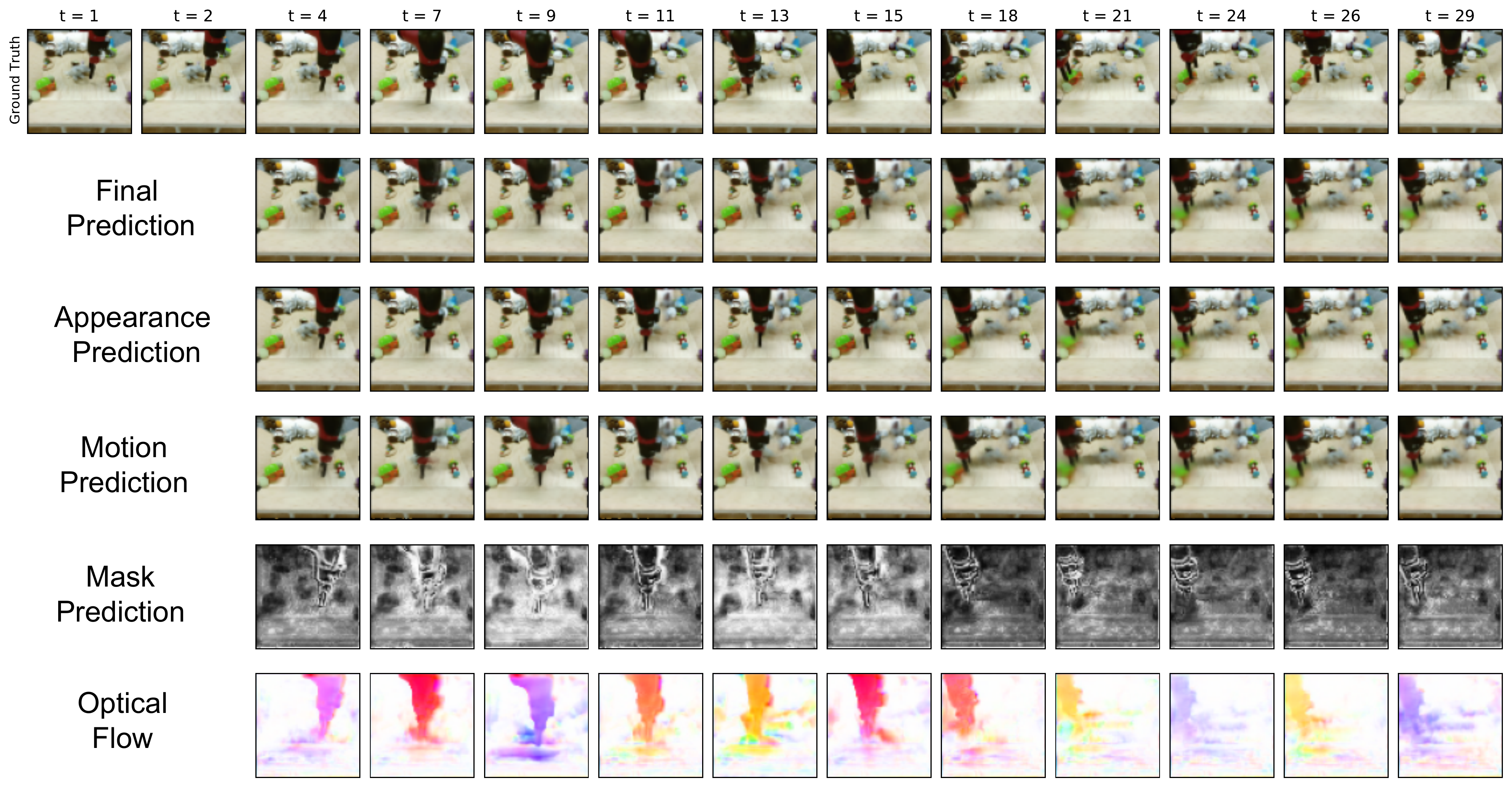}
\caption{\textbf{Results on BAIR.} Optical flow consistently changes from one frame to the next,
showing that our model can learn the dataset dynamics. Note that predicted flow is from the target frame to the current frame since we apply inverse warping.}
\label{fig:bair_ex}
\end{figure}
\begin{figure}[H]
\centering
\includegraphics[width=\textwidth]{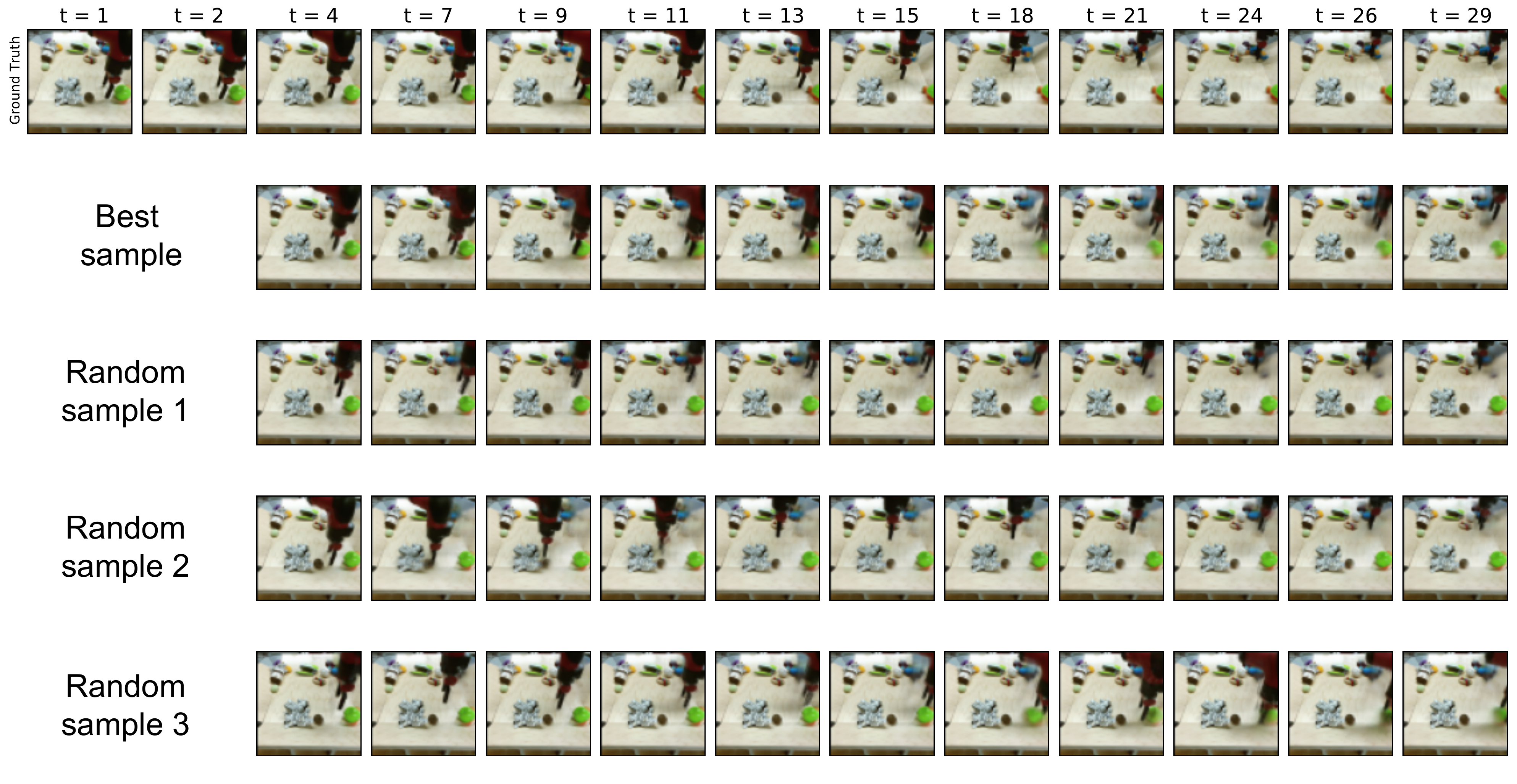}
\caption{\textbf{Random Samples.} We show the best sample and three random samples generated. Random samples look very similar in the beginning due to regular motion in the conditioning frames. Towards the end of the sequence, samples start looking different which shows that our model can generate diverse results.}
\label{fig:bair_random}
\end{figure}

\begin{figure}[H]
\centering
\includegraphics[width=\textwidth]{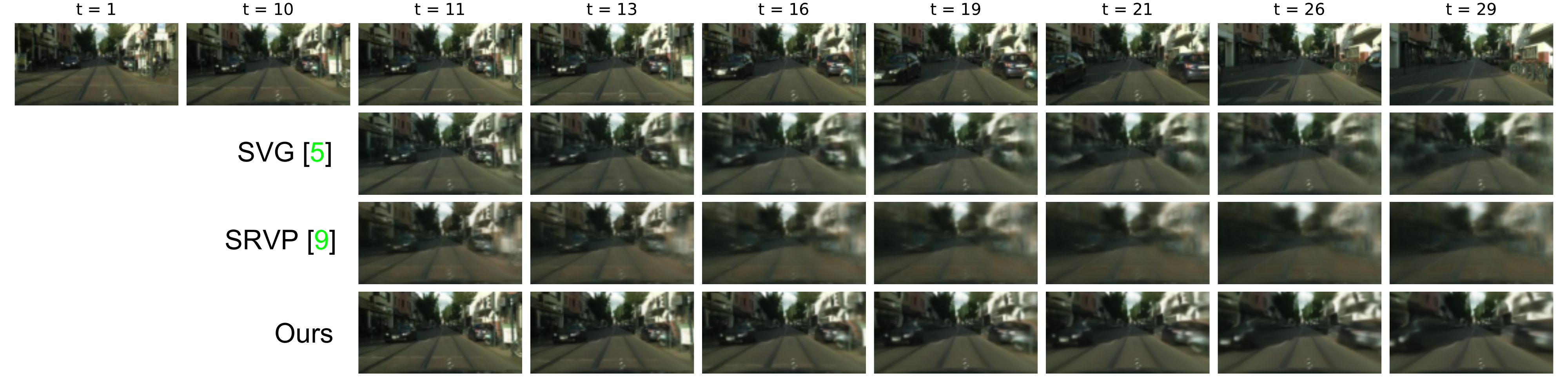}
\caption{\textbf{Full sequence of Cityscapes in \figref{fig:teaser}} We show the full sequence comparisons with baseline method, SVG~\cite{Denton2018ICML}, and state-of-the-art method, SRVP~\cite{Franceschi2020ICML}. Our model can model the ego-motion while both SRVP and SVG suffers from it significantly.}
\label{fig:full_seq_teaser_city}
\end{figure}

\begin{figure}[H]
\centering
\includegraphics[width=\textwidth]{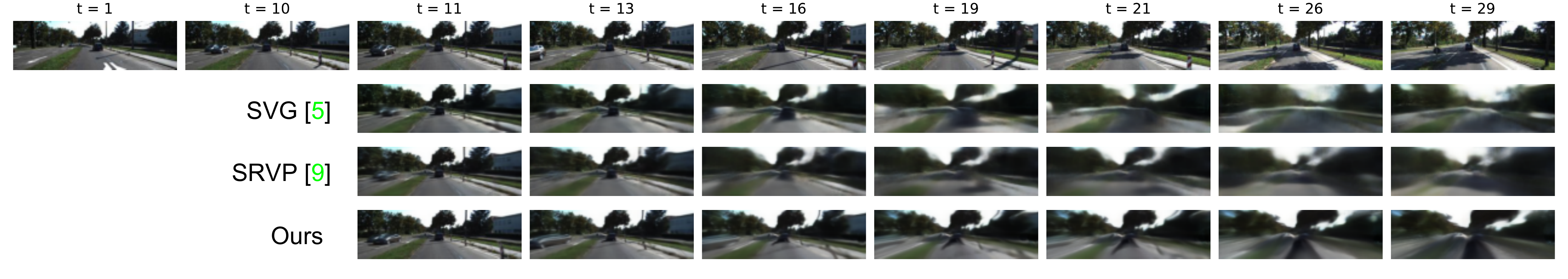}
\caption{\textbf{Full sequence of KITTI in \figref{fig:teaser}} We show the full sequence comparisons with baseline method, SVG~\cite{Denton2018ICML}, and state-of-the-art method, SRVP~\cite{Franceschi2020ICML}. Our model can model both the ego-motion and independently moving objects while both SRVP and SVG cannot reconstruct the future frames successfully.}
\label{fig:full_seq_teaser_kitti}
\end{figure}

\end{appendix}

\end{document}